%% file: manuscript.tex
\documentclass[lettersize,journal]{IEEEtran}
\IEEEoverridecommandlockouts
\usepackage{adjustbox}
\usepackage{cite}
\usepackage{amsmath,amssymb,amsfonts}
\usepackage{algorithmic}
\usepackage{graphicx,subcaption}
\usepackage{textcomp}
\usepackage{multirow}
\usepackage{pifont}
\usepackage{makecell}
\usepackage{stfloats} 
\usepackage{array}
\usepackage[a4paper, total={184mm,239mm}]{geometry}
\usepackage{url}
\makeatletter
\AtBeginDocument{\DeclareMathVersion{bold}
\SetSymbolFont{operators}{bold}{T1}{times}{b}{n}
\SetMathAlphabet{\mathrm}{bold}{T1}{times}{b}{n}
\SetMathAlphabet{\mathit}{bold}{T1}{times}{b}{it}
\SetMathAlphabet{\mathbf}{bold}{T1}{times}{b}{n}
\SetMathAlphabet{\mathtt}{bold}{OT1}{pcr}{b}{n}
\SetSymbolFont{symbols}{bold}{OMS}{cmsy}{b}{n}
\renewcommand\boldmath{\@nomath\boldmath\mathversion{bold}}}
\makeatother

\def\BibTeX{{\rm B\kern-.05em{\sc i\kern-.025em b}\kern-.08em
    T\kern-.1667em\lower.7ex\hbox{E}\kern-.125emX}}

\usepackage{fancyhdr}
\usepackage{threeparttable} 
\usepackage{color}
\usepackage{soul}

\fancypagestyle{firstpage}{%
 \fancyhf{}
 
 \lhead{This work has been accepted in an IEEE journal. DOI: \url{10.1109/ACCESS.2026.3707342}}
 }

\begin{document}

\newcommand{\MT}[1]{{\color{red}MT: #1}}
\newcommand{\GA}[1]{{\color{black}GA: #1}}

\newcommand{\yellowstar}{\textcolor[rgb]{1.0,0.647,0.0}{$\bigstar$}}
\newcommand{\blackdot}{%
  {\textcolor[rgb]{0,0,0}{\ding{108}}}%
}
\newcommand{\reddot}{%
  {\textcolor[rgb]{1,0,0}{\ding{108}}}%
}

\newcommand{\greendot}{%
  {\textcolor[rgb]{0.176,0.627,0.176}{\ding{108}}}%
}

\newcommand{\drawcross}{%
  {\textcolor[rgb]{0.87,0.32,0.33}{\Large×}}%
}

\title{PERTINENCE: Input-based Opportunistic Neural Network Dynamic Execution} 
\author{
Omkar Shende$^1$, Gayathri Ananthanarayanan$^2$, Marcello Traiola$^3$\\
$^{1,2}$Department of Computer Science and Engineering, Indian Institute of Technology Dharwad, Karnataka, India\\$^3$University of Rennes, CNRS, Inria, IRISA, Rennes, France\\
$^1$212011004@iitdh.ac.in, $^2$gayathri@iitdh.ac.in, $^3$marcello.traiola@inria.fr}

\maketitle

\begin{abstract}
Deep neural networks (DNNs) have become ubiquitous thanks to their remarkable ability to model complex patterns across various domains such as computer vision, speech recognition, robotics, etc. While large DNN models are often more accurate than simpler, lightweight models, they are also resource- and energy-hungry. Hence, it is imperative to design methods to reduce reliance on such large models without significant degradation in output accuracy. The high computational cost of these models is often necessary only for a reduced set of challenging inputs, while lighter models can handle most simple ones. Thus, carefully combining properties of existing DNN models in a dynamic and input-based way opens opportunities to improve efficiency without compromising accuracy.
In this work, we introduce PERTINENCE, a novel runtime method to
dynamically select, from a set of pre-trained models, the most lightweight one that correctly processes a given input. 
To achieve this, we use an ML-based input dispatcher that chooses the model to process the current input.
We explore the training space of the input dispatcher using a genetic algorithm to identify the Pareto front of training solutions that offer trade-offs between overall accuracy and computational efficiency.
We showcase our approach on state-of-the-art Convolutional Neural Networks (CNNs) trained on the CIFAR-10 and CIFAR-100 datasets, as well as Vision Transformers (ViTs) trained on the TinyImageNet dataset. Furthermore, we showcase PERTINENCE in a YOLO-based road occupancy estimation application that processes real-time video feeds from cameras installed at the road’s intersection. We report results showing PERTINENCE's ability to provide alternative solutions to existing state-of-the-art models, balancing accuracy and the number of operations. The results show that, thanks to dynamic selection of pre-trained models, PERTINENCE achieves comparable or higher accuracy compared to the best pre-trained model, with up to 36\% fewer operations with equivalent or lower end-to-end inference time using tunable invocation intervals.
\end{abstract}


\thispagestyle{firstpage}

\section{Introduction}
\label{sec:intro}
Large and complex Deep Neural Networks (DNNs) have traditionally been used to achieve high prediction accuracy. However, the computational cost overhead of these models often outweighs the marginal gains in accuracy they offer compared to lightweight variants. As we shift gears towards carbon-aware, sustainable computing, it is imperative that we develop new techniques to identify opportunities for employing smaller, simpler, and lightweight models that achieve comparable performance to larger models, albeit with significantly lower computational costs. To this end, in this work, we leverage the interesting fact that, during inference, not all inputs require large, complex models to be processed correctly~\cite{hapi}; instead, different inputs may require varying levels of model complexity.

Reducing the computational cost of CNN inference has attracted significant interest due to the growing demand for deploying DNN models in resource-constrained environments, such as mobile devices, embedded systems, and edge computing platforms. Key approaches proposed in the existing literature to address this include model compression techniques such as pruning, quantization, knowledge distillation ~\cite{model_compression,pruning}, and weight sharing~\cite{weight_share}. Another relevant area of work in this direction is neural architecture search (NAS)~\cite{nas}, which focuses on the automated design of CNNs optimized for performance, accuracy, and other design criteria, such as memory footprint. A large body of work exists in these areas and has been effectively summarized in these surveys~\cite{nas2, nas, pruning}. 
While these strategies aim to reduce computational costs, they are largely used at design time. In other words, they aim to reduce the model's complexity at training or deployment time without changing the computational behavior during inference.
In contrast, our work focuses on dynamically adapting computational load in real time to different inputs. Some earlier works~\cite{mcdnn, mobisr} have proposed solutions to improve the performance of  applications such as super-resolution and video processing by adapting the computation to the inputs. Han et al.~\cite{dynNN-survey}  provide a comprehensive survey of dynamic neural networks, categorizing them into instance-wise dynamic models, spatial-wise adaptive networks, 
and temporal-wise dynamic models for sequential data, all sharing the fundamental motivation of avoiding uniform computational allocation across inputs of varying complexity, which similarly underlies the design proposed in this work.
Moreover, techniques such as early exit networks~\cite{earlyexit,msdnet,hapi} and slimmable neural networks~\cite{slim1,slim2} align with this objective, as they allow a model to terminate inference early when it reaches a prediction with sufficient confidence, eliminating the need to process all network layers for every input. 
Early exit networks~\cite{earlyexit-extra} are built around a backbone architecture with additional exit heads (or classifiers) at various depths. During inference, the sample traverses the network, moving sequentially through the backbone and each exit. Training early-exit networks is inherently challenging, as it requires more than simply training the backbone and exits independently or with basic initialization. The training sequence needs to be carefully planned so that each exit learns to make predictions based on the appropriate features available at that processing stage. This requires a coordinated approach in which the training of the exits aligns with the backbone's capabilities. 
Similarly, slimmable networks are a class of networks that allow runtime selection of model width by changing the number of channels in each layer. Slimming creates submodels with latency-accuracy trade-offs. 
Another class of methods includes Mixture of Experts (MoE) architectures~\cite{moe}, which consist of an ensemble of neural networks, each acting as an “expert” in a specific domain within a larger problem. These architectures designate multiple ``experts", each serving as a sub-network within the overall framework. They also train a gating network (or router)~\cite{moe1} that selectively activates the expert(s) best suited for a given input.

Despite the remarkable progress in DNN optimization, existing approaches largely treat all inputs uniformly, applying the same  computational budget regardless of the inherent difficulty of each 
input. This is fundamentally suboptimal: a simple and clear image of a single object requires far less computation to classify correctly than a cluttered, low-resolution scene. Furthermore, approaches such 
as model compression and NAS operate at design time and do not adapt to input characteristics during inference. Early exit networks and slimmable networks, while runtime-adaptive, require careful joint 
training procedures and architectural modifications, limiting their applicability to existing pre-trained models. This motivates the need  for a runtime approach that leverages existing pre-trained models of varying complexity and dynamically adapts the computational effort based on the input, without any retraining or architectural modifications.

To the best of our knowledge, there is little research on combining at runtime the capabilities of different existing pre-trained predictors with different accuracy-cost trade-offs by dynamically dispatching inputs to the most lightweight one able to correctly process them. 
This approach would enable combining the properties of existing NN models (e.g., high accuracy or efficient execution) to achieve hybrid synergy.

Therefore, we propose \textit{PERTINENCE}, a new runtime method that intelligently combines pre-trained DNN models at runtime to reduce computational effort.

We show that by dynamically combining state-of-the-art models, we can achieve better trade-offs between accuracy and the number of operations than when using the models individually.
We draw inspiration from the MoE approach, particularly in the design of the input dispatcher. However, PERTINENCE operates along a fundamentally different axis. Traditional MoE methods typically split the dataset upfront, train separate expert models for each partition, and use a learned gating mechanism to dispatch inputs during inference. In contrast, PERTINENCE does not need a dataset split and expert training. Rather, it learns to determine which pre-trained model to use at runtime for achieving a better trade-off between accuracy and computational complexity.
Unlike MoE systems, which often require retraining multiple expert models as dataset complexity increases, PERTINENCE leverages existing pre-trained networks, avoiding re-training. Additionally, MoE frameworks primarily focus on improving accuracy, whereas PERTINENCE explicitly targets minimizing operations to achieve computational efficiency while maintaining high accuracy.
We perform an evolutionary exploration of the input dispatcher training while considering accuracy, FLOPS, power consumption, and inference time. We also study the impact of varying the dispatcher rate in a YOLO-based real-time video feed application. 
The experimental results show that PERTINENCE provides superior solutions, achieving higher accuracy at reduced power consumption without impacting inference time.

In summary, we propose the following contributions:
\begin{itemize}
\item We introduce PERTINENCE, a runtime method for selecting the least resource-intensive pre-trained model that correctly processes a given input, using an ML-based input dispatcher.
\item We explore the input-dispatcher training space using an evolutionary approach.
\item We conduct an extensive experimental campaign on state-of-the-art Neural Networks, including CNNs and Vision Transformers. Finally, we apply PERTINENCE on a YOLO-based real-time video feed application.
\end{itemize}

\section{Proposed methodology} 
\begin{figure}[t]

    \centering
\includegraphics[width=0.9\columnwidth]{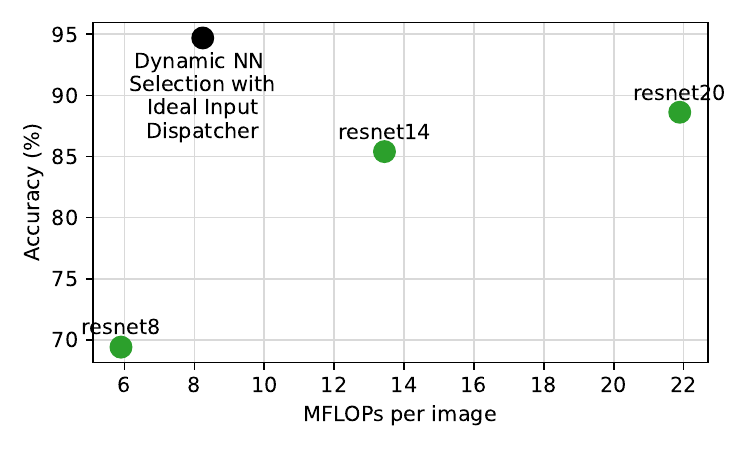}
    \caption{MFLOPs- Accuracy tradeoff for three different CNN models trained on CIFAR-10 dataset compared with using an ideal Input Dispatcher to use the three CNNs depending on the input opportunistically.}
    \label{fig:mot1}
\end{figure}

In this section, we present the problem formulation and the details related to the proposed methodology. Let us start with a motivating example to introduce the concept. Consider three different CNN models \texttt{resnet8}, \texttt{resnet14} and \texttt{resnet20} trained on CIFAR-10 dataset~\cite{cifar_data}. Fig.~\ref{fig:mot1} illustrates the trade-offs between computational cost (measured in MFLOPs) and accuracy for these three models. \texttt{resnet8} model incurs a computational cost of 5.9 MFLOPs for a single image inference and achieves an accuracy of 69.4\% on the CIFAR-10 test dataset.
\texttt{resnet20} and \texttt{resnet14} require $3.7\times$ and $2.3\times$ more computations, respectively, compared to \texttt{resnet8}, while achieving accuracy improvements of 19.2\% and 16\%, respectively. 
Table~\ref{tab:mot1} shows the distribution of CIFAR-10 test dataset (10000) images correctly classified by different combinations of three CNN models. 

The second-to-last row clearly shows that only 527 of 10000 images require the larger model (\texttt{resnet20}) for correct classification, whereas for 6947 images, the smallest model (\texttt{resnet8}) alone would be sufficient.

\begin{table}[b]
\caption{Distribution of CIFAR-10 test dataset (10000) images correctly classified by different combinations of three CNN models.}
    \centering
    \begin{tabular}{c|c|c|c}
     \textbf{\texttt{resnet8}} & \textbf{\texttt{resnet14}} & \textbf{\texttt{resnet20}} & \textbf{\#Images} \\
     \hline
     \ding{51} &\ding{51} &\ding{51} &6320\\
     \hline
     \ding{51} &\ding{51} &\ding{55}&228\\
     \hline
     \ding{51} &\ding{55}&\ding{51} &256\\
     \hline
     \ding{51} &\ding{55}&\ding{55}&143\\
     \hline
     \ding{55}&\ding{51} &\ding{51} &1759\\
     \hline
     \ding{55}&\ding{51} &\ding{55}&237\\
     \hline
     \ding{55}&\ding{55}&\ding{51} &527\\
     \hline
     \ding{55}&\ding{55}&\ding{55}&530\\

    \end{tabular}

    \label{tab:mot1}

\end{table}

It becomes clear the advantage that would bring an \textit{ideal Input Dispatcher} dynamically predicting which model to use for a particular input, to produce a correct prediction and minimize the needed floating point operations (FLOPs). For the three CNN models used in our example, using the image distribution data from Table~\ref{tab:mot1}, the ideal Input Dispatcher will select \texttt{resnet8} to be used for 6947 images, \texttt{resnet14} for 1996 images and \texttt{resnet20} for 527 images. There are 530 images that none of the CNN models can correctly classify; in such cases, let us assume that the ideal predictor would choose \texttt{resnet8} to minimize the computational cost associated with the incorrectly classified images.  
By correctly dispatching the inputs to the most suitable network, a 62.11\% reduction in MFLOPs and an increase of 6.08\% in accuracy can be achieved, compared to using the most accurate network (\texttt{resnet20}, in this case) for all inputs, as also shown in Fig.~\ref{fig:mot1}. 
Of course, such an ideal input dispatcher is not implementable with zero cost. We formally define the addressed problem below.

\subsection*{Problem Formulation}

Let $\mathcal{M} = \{M_1, M_2, \dots, M_N\}$ be a set of $N$ pre-trained 
DNN models, ordered such that $M_1$ is the least computationally expensive 
and least accurate, and $M_N$ is the most computationally expensive and most 
accurate. Each model $M_i$ is characterized by two properties: its inference 
accuracy $\alpha_i$ and its computational cost $\phi_i$, measured as a function of the 
Multiply-Accumulate Operations and is represented by MFLOPs. The models in $\mathcal{M}$ are 
selected from the Pareto front of the accuracy-vs-computational cost space, ensuring 
that no model is strictly dominated by another in both objectives 
simultaneously.

\bgroup\color{black}
Given an input sample $x$ drawn from an input space $\mathcal{X}$, let 
$\mathcal{C}(M_i, x) \in \{0, 1\}$ denote whether model $M_i$ correctly 
processes input $x$, where $\mathcal{C}(M_i, x) = 1$ indicates a correct 
prediction and $\mathcal{C}(M_i, x) = 0$ indicates an incorrect one. The 
goal is to learn a dispatcher function $D_{p}$:

\begin{equation}
    D_{p}: \mathcal{X} \rightarrow \mathcal{M}
    \label{eq:dispatcher}
\end{equation}

\noindent such that for each input $x$, $D_{p}(x)$ selects the least 
computationally expensive model $M_i \in \mathcal{M}$ that correctly 
processes $x$, i.e.:

\begin{equation}
    D_{p}(x) = M_i, \quad \text{where } i = \arg\min_j \{ \phi_j \mid \mathcal{C}(M_j, x) 
    = 1 \}
    \label{eq:dispatch_rule}
\end{equation}

This formulation defines a bi-objective optimization problem. Given a dataset 
$\mathcal{X}$ composed of $|\mathcal{X}|$ input samples, the two competing objectives are:

\begin{equation}
    \text{Maximize:} \quad \alpha_{sys} = \frac{1}{|\mathcal{X}|} \sum_{k=1}^{|\mathcal{X}|} 
    \mathcal{C}(D_{p}(x_k), x_k)
    \label{eq:obj_accuracy}
\end{equation}

\begin{equation}
    \text{Minimize:} \quad \phi_{sys} = \frac{1}{|\mathcal{X}|} \sum_{k=1}^{|\mathcal{X}|} 
    \phi_{D_{p}(x_k)}
    \label{eq:obj_flops}
\end{equation}
\egroup
\bgroup\color{black}
\noindent where $\alpha_{sys}$ is the overall system accuracy and $\phi_{sys}$ 
is the average number of operations per input sample, considering all 
operations performed by both the dispatcher $D_{p}$ and the selected model 
$D_{p}(x_k)$ for each input $x_k$.

The problem is subject to the following constraints:
\begin{itemize}
    \item All models in $\mathcal{M}$ are pre-trained and fixed; no 
    retraining or fine-tuning of the backbone DNN models is performed.
    \item The dispatcher $D_{p}$ is implemented as a lightweight ML model 
    consisting of a pre-trained feature extractor and a single to-be-trained fully 
    connected (FC) layer, ensuring that its computational overhead is 
    small relative to the models in $\mathcal{M}$.
    \item The dispatcher operates entirely at runtime, making model 
    selection decisions based solely on the input $x$, without any 
    prior knowledge of the correct prediction label for $x$.
\end{itemize}

Since the two objectives in Eq.~\eqref{eq:obj_accuracy} and 
Eq.~\eqref{eq:obj_flops} are competing, routing all inputs to $M_N$ 
maximizes accuracy but also maximizes computational cost, while routing all inputs to $M_1$ minimizes cost but degrades accuracy. Hence, we explore the training space of the dispatcher to obtain solutions offering different tradeoffs between the two objectives. To do so, as discussed in Section~\ref{sec:input-dispatcher}, we use penalties in the loss function while training the FC layer of the dispatcher to ‘steer' it towards more accurate or less computationally expensive solutions. Since the penalties are defined on a continuous range, finding exact solutions of this Multi-Objective Optimization Problem (MOP) is computationally 
intractable. Therefore, as described in the 
following sections, we resort to a Multi-objective Evolutionary Algorithm 
(MOEA), specifically NSGA-II~\cite{NSGA-II}, to approximate the 
Pareto front of dispatcher training configurations, each offering a 
different trade-off between $\alpha_{sys}$ and $\phi_{sys}$.
\egroup
\begin{figure}[b]
    \centering
    \includegraphics[width=.9\columnwidth]{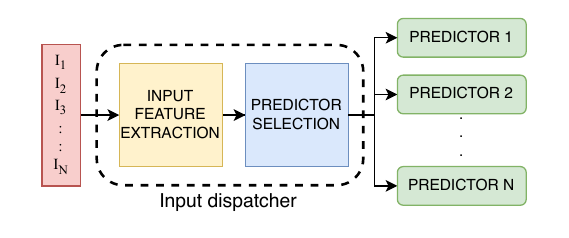}
    \caption{PERTINENCE Approach}
    \label{fig:gen-method}
\end{figure}
In this work, we focus on an input dispatcher that dynamically assigns the inputs to the most appropriate predictor for performing the task, as depicted in Fig.~\ref{fig:gen-method}. Depending on the desired objectives, different dynamic dispatch choices can be made. 
Identifying the optimal choice is far from trivial, especially when faced with conflicting objectives such as accuracy and computational cost. To effectively balance competing objectives, the selection method should incorporate a mechanism to predict the impact of each choice across multiple objectives, determining the most suitable predictor for the given input.

The approach is composed of three steps:
\begin{enumerate}
    \item choose an ML task (e.g., image classification) and the competing objectives determining the space to explore (e.g., accuracy and number of operations);
    \item in the above space, select the existing state-of-the-art (SOTA) predictors lying on the \textit{Pareto-front} (e.g., the NNs offering the best trade-offs between accuracy and number of operations for a given task);
    \item create an input dispatcher $D_{p}$ to opportunistically select at run-time the most suitable predictor depending on the input;
\end{enumerate}

\begin{table}[htp]
\color{black}
\caption{Nomenclature}
\label{tab:nomenclature}
\renewcommand{\arraystretch}{1.3}

\begin{tabular}{p{2cm} p{6cm}}
\hline
\multicolumn{2}{l}{\textbf{Abbreviations and Acronyms}} \\
\hline
DNN     & Deep Neural Network \\
CNN     & Convolutional Neural Network \\
ViT     & Vision Transformer \\
NAS     & Neural Architecture Search \\
MoE     & Mixture of Experts \\
SOTA    & State of the Art \\
FC      & Fully Connected \\
\hline
\multicolumn{2}{l}{\textbf{Optimization and Algorithmic Terms}} \\
\hline
MOP     & Multi-objective Optimization Problem \\
MOEA    & Multi-objective Evolutionary Algorithm \\
NSGA-II & Non-dominated Sorting Genetic Algorithm II \\
GA      & Genetic Algorithm \\
SBX     & Simulated Binary Crossover \\
INS     & Inverse of Number of Samples \\
ISNS    & Inverse of Square Root of Number of Samples \\
ENS     & Effective Number of Samples \\
\hline
\multicolumn{2}{l}{\textbf{Metrics}} \\
\hline
FLOP    & Floating Point Operation \\
MFLOP   & Mega Floating Point Operation \\
MAC     & Multiply-Accumulate Operation \\
FPS     & Frames Per Second \\
\hline
\multicolumn{2}{l}{\textbf{Application Terms}} \\
\hline
YOLO    & You Only Look Once \\
SoC     & System on Chip \\
FPS     & Frames Per Second \\
\hline
\end{tabular}

\vspace{0.5cm}

\begin{tabular}{p{2cm} p{6cm}}
\hline
\multicolumn{2}{l}{\textbf{Mathematical Symbols}} \\
\hline
\multicolumn{2}{l}{\textit{Problem Formulation (Section II)}} \\
\hline
$\mathcal{M}$           & Set of $N$ pre-trained DNN models \\
$M_i$                   & $i$-th pre-trained DNN model \\
$N$                     & Total number of pre-trained DNN models 
                          in $\mathcal{M}$ \\
$\alpha_i$              & Inference accuracy of model $M_i$ \\
$\phi_i$                & Computational cost of model $M_i$ 
                          (in MFLOPs) \\
$\mathcal{X}$           & Input space \\
$x$                     & An input sample drawn from $\mathcal{X}$ \\
$\mathcal{C}(M_i, x)$   & Correctness function; equals 1 if model 
                          $M_i$ correctly processes input $x$, 
                          0 otherwise \\
$D_p(\cdot)$              & Dispatcher function 
                          $D_p: \mathcal{X} \rightarrow \mathcal{M}$ \\
$\alpha_{sys}$          & Overall system accuracy across all 
                          input samples \\
$\phi_{sys}$            & Average number of operations per input 
                          sample across the full pipeline \\
\hline
\multicolumn{2}{l}{\textit{Loss Function (Section II)}} \\
\hline
$|\mathcal{X}|$                     & Total number of input samples in 
                          the dataset \\
$k$                     & Index of the $k$-th input sample \\
$y_{true,k}$            & True class label of the $k$-th sample \\
$y_{pred,k}$            & Predicted class label of the 
                          $k$-th sample \\
$L_k$                   & Per-sample loss for the $k$-th sample \\
$L_{base,k}$            & Base cross-entropy loss for the 
                          $k$-th sample \\
$L$                     & Overall loss across all $M$ samples \\
$P$                     & Penalty matrix of size $N \times N$ \\
$P[i,j]$                & Penalty applied for misclassifying 
                          class $i$ as class $j$ \\
\hline
\multicolumn{2}{l}{\textit{Dynamic Dispatching (Section IV)}} \\
\hline
$D$                     & Dispatcher invocation interval 
                          (in frames) \\
$\theta$                & Composite change score computed by 
                          the change detector \\
$t$                     & Threshold for composite change 
                          score $\theta$ \\
\hline
\end{tabular}

\end{table}

In the remainder of the paper, we apply the above methodology to the practical case of image classification using DNNs. In this work, we use the term DNNs to refer specifically to convolutional neural networks (CNNs) and vision transformers (ViTs).  We showcase the proposed approach with CNNs and ViTs. The Table~\ref{tab:nomenclature} provides the details of all the symbols and terms used in this work.

\subsection{Input Dispatcher for DNN-based image classification}\label{sec:input-dispatcher}
We want to be able to dynamically predict the best DNN model for image classification of a given dataset, based on the input images themselves. In this work, we consider two competing objectives, i.e., accuracy and number of operations (expressed as MFLOPS). In the so-defined space, we select publicly available state-of-the-art DNN models, achieving the best trade-offs between accuracy and number of operations.
Since the two selected objectives are competing, different dispatching choices lead to different trade-offs between the objectives.

\begin{figure}[h!]
    \centering
    \includegraphics[trim={15pt 10pt 15pt 10pt},clip,width=\columnwidth]{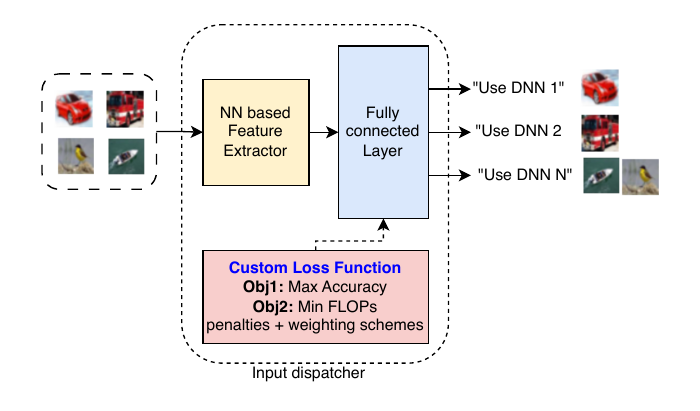}
    \caption{Proposed Input Dispatcher Design}
    \label{fig:bin-pred}
\end{figure}

For the input dispatcher, we use a neural network-based approach to extract meaningful features from the input image.
Then, to select the DNN to perform the inference depending on the input image, we train \textit{a single fully connected layer} on the extracted features to propose the use of a specific DNN among the ones selected (from the \textit{Pareto-front} in the selected space). Fig.~\ref{fig:bin-pred} sketches the design methodology for such a dispatcher.
The \textit{fully connected} (FC) layer \textit{input size} and \textit{number of output classes} will depend on the last layer of the feature extractor network and on the number of SOTA DNNs to dispatch inputs to (i.e., the green boxes in Fig.~\ref{fig:gen-method}), respectively.

The Pareto front includes different CNNs, i.e., small, lightweight, less-accurate models, up to large, computing-intensive, more accurate ones. 
In Figure~\ref{fig:DNN-pareto}, we consider a set of SOTA CNNs used for image classification on different datasets, i.e. CIFAR-10, CIFAR-100~\cite{cifar_data}, and Tiny Imagenet~\cite{tinyimage_data}. For the first two, we consider SOTA CNNs that lie on the defined Pareto front, and for the last one, SOTA ViTs that lie on the defined Pareto front.
The goal of the approach is to dynamically select the most suitable DNN to process the image, based on the desired trade-off. In our exploration we focus on \textit{maximizing accuracy while, at the same time, minimizing the number of operations required to perform the inferences}.

\input{pareto-sota-dnns}

As it can be easily imagined, the problem we are tackling suffers from the class imbalance problem. To show the phenomenon, let us consider the CNNs in Fig.~\ref{fig:cf10-pareto} for the CIFAR-10 datasets,~\cite{cifar_data,cifar_models}. We computed the number of operations per image and the accuracy over the test set. The accuracy difference between the least and most accurate models is 25.6\%. This means that for $\approx$70\% of input samples, the least accurate -- and smallest -- model will be sufficient, and $\approx$ 25\% of samples will require larger models. This shows that many input images will be labeled to be managed by the smallest model (majority class), while the other classes will have significantly fewer samples (minority classes), leading to the class imbalance problem. Table~\ref{tab:sample_weights} reports the sample distribution statistics for CIFAR-10 training dataset for the case four of the CNN models on the Pareto front.
\begin{table}[htp]
    \caption{Input Sample distribution across various Classes }
    \centering
    \begin{tabular}{c|c|c}
    \textbf{Model}   & \textbf{\% of Samples} & \textbf{Class} \\
    \hline
      \texttt{resnet8}   & 68.46\% & Majority\\
      \hline
      \texttt{resnet14} & 20.40\% & Minority 1\\
      \hline
      \texttt{shufflenetv2\_x0\_5}  & 9.47\% & Minority 2\\
      \hline
      \texttt{vgg16\_bn} & 1.67\% & Minority 3\\
    \end{tabular}
    \label{tab:sample_weights}
\end{table}

To address this, we use sample weights in the loss function while training the FC layer. This is done to assign different weights to the loss of each sample, depending on whether it belongs to the majority or minority classes. Specifically, we assign a higher weight to the loss of samples from minority classes. In this work, we explored three weighting schemes to compute sample weights: the inverse of the number of samples (INS), the inverse of the square root of the number of samples (ISNS), and the Effective Number of Samples (ENS)~\cite{ens_cvpr19} weighting scheme.

\textit{Computing Penalties:} Typically, NNs are trained to maximize accuracy. In this work, we address a bi-objective problem, aiming to maximize accuracy while minimizing the computational effort required for inference. To achieve this balance, we introduce penalties into the loss function of the trained FC layer.  As an example case, let us assume that three DNN models are selected from the Pareto front for inference, categorized as follows: class 0 (highest accuracy, requiring a lare number of computations), class 1 (medium accuracy, requiring a moderate number of computations), and class 2 (low accuracy, requiring fewer computations). Misclassifying class 0 as class 1 or 2 reduces the overall classification accuracy, while misclassifying class 2 as either class 0 or class 1 increases computational costs but does not affect accuracy. Depending on the final goal (i.e., desired trade-off between accuracy and number of operations) different values for the penalty matrix have to be assigned. For the above example, the penalty matrix is defined in Eq.~\eqref{eq:penalty_matrix} as:
\begin{equation}
P =
\begin{bmatrix}
0 & p_{01} & p_{02} \\
p_{10} & 0 & p_{12} \\
p_{20} & p_{21} & 0
\end{bmatrix}
\label{eq:penalty_matrix}
\end{equation}

The diagonal elements \( P[i, i] \) are \( 0 \), i.e., no loss contribution for correct predictions. The off-diagonal elements \( P[i, j] \) represent penalties applied for specific misclassifications. The penalty matrix \( P \) in Eq.~\eqref{eq:penalty_matrix} modifies the base cross-entropy loss for each prediction.

\textit{Custom Loss Function:} For each input sample \( k \), the loss is computed as follows:
\begin{equation}
    L_k =
\begin{cases} 
0, & \text{if } y_{\text{true}, k} = y_{\text{pred}, k} \\ 
L_{\text{base}, k} \cdot P[y_{\text{true}, k}, y_{\text{pred}, k}], & \text{if } y_{\text{true}, k} \neq y_{\text{pred}, k}
\end{cases}
\label{eq:sample_loss}
\end{equation}

where  \( y_{\text{true}, k} \) is the true class of the \( k \)$^{th}$ sample, 
\( y_{\text{pred}, k} \) is the predicted class of the \( k \)$^{th}$ sample, \( L_{\text{base}, k} = \text{CrossEntropy}(y_{\text{true}, k}, y_{\text{pred}, k})\) is the base cross-entropy loss.  The per-sample loss is defined in Eq.~\eqref{eq:sample_loss} and the 
overall loss across all $M$ images in the input data set is computed as shown in 
Eq.~\eqref{eq:total_loss}.
\begin{equation}
L = \frac{1}{M} \sum_{k=1}^{M} L_k
\label{eq:total_loss}
\end{equation}

\begin{figure*}
    \centering
    \includegraphics[trim={20pt 15pt 20pt 15pt},clip,width=0.9\textwidth]{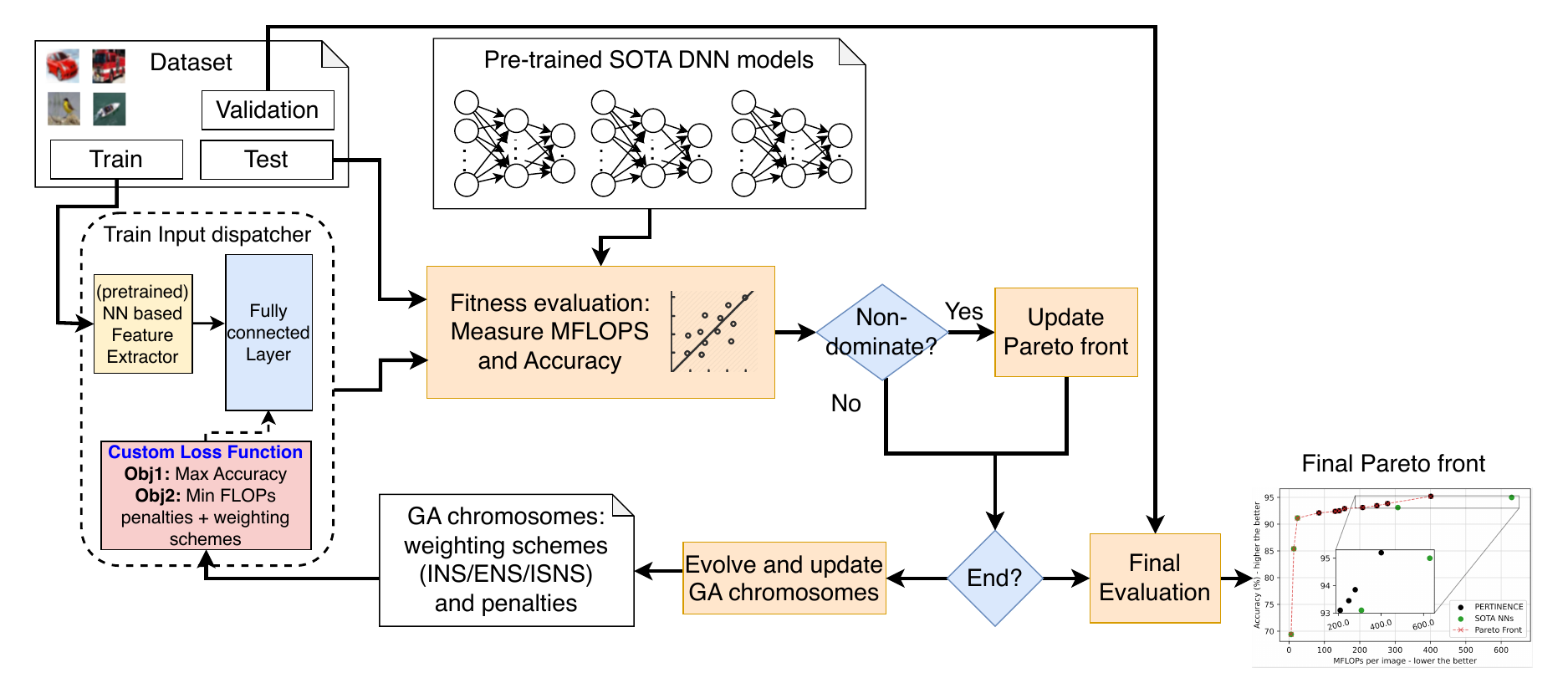}
    \caption{Flow of the proposed PERTINENCE approach}
    \label{fig:flow}
\end{figure*}
To explore the space of possible trade-offs between the number of operations (MFLOPS) and accuracy, and compare these with the SOTA DNNs, we resort to a genetic algorithm to explore different combinations of penalties and weighting schemes. 
We model this problem as a Multi-objective Optimization Problem (MOP).
Basically, a MOP consists of a set of~\textit{objective functions} to be either minimized or maximized subject to a set of constraints.
Since different objectives often represent conflicting objectives, the exploration goal is to seek for a set of equally good solutions being close to the \textit{Pareto front}.
Given two solutions $x,y : x \ne y$, $x$ is said to \textit{dominate} $y$ iff $x$ is better or equally good in all objectives than $y$ and at least better in one objective.
If a solution is not dominated by any others, it is called a \textit{Pareto-optimal} solution.
In the case of complex MOP, exact resolution algorithms turn out to be too computationally expensive. Therefore, usually they are not applicable when the search space is very large. Consequently, we resort to a Multi-objective evolutionary algorithm (MOEA) heuristic to produce an approximation of the Pareto front in a rather reasonable time. These are largely used in the literature to find Pareto fronts for MOP~\cite{deb2001multi, NSGA-II, van2000multiobjective}. 
MOEAs operate on a set of \textit{individuals}, called \textit{population}, that evolves and, eventually, converges to a set of Pareto-optimal solutions.
Each individual is represented as a \textit{chromosome}, i.e., a data structure encoding the search space.
During the evolution process, new offsprings are generated either through or in combination of {\em mutation} and {\em crossover}.
A \textit{crossover} takes two parent chromosomes to produce a new chromosome.
Applied to our problem, we consider total FLOPS and total accuracy as the two competing objectives, the former to be minimized and the latter to be maximized.
An individual of the MOEA corresponds to a FC layer training, i.e., with different weighting scheme and loss function penalties. Depending on the training, the FC layer will dispatch the inputs to the subsequent DNN models differently, leading to different trade-offs between our objectives.
In particular, we resorted to NSGA-II~\cite{NSGA-II} and the number of chromosomes that we use to encode the search space depends on the number of networks after the input dispatcher. Indeed, we need $(\textit{num. of DNNs})^2$ chromosomes to encode the $P$ matrix and an extra one for the weighting scheme. For example, when dispatching to 3 DNNs, 10 chromosomes are required in the exploration.
In this way, by mutating the chromosomes, the MOEA makes the PERTINENCE variants evolve towards the Pareto front.

Fig.~\ref{fig:flow} sketches the flow of the proposed exploration. We start by generating random chromosomes (i.e., penalties and a weighting scheme) and assign those to an initial population. 
Then, for each individual, we train the FC layer using the custom loss function (including penalties and a weighting scheme), and, as input, the features extracted from the training set images. To extract these features, we use a pre-trained feature extractor (i.e., the backbone of an SOTA DNN). We relabeled the images so that the FC layer assigns each to the least computationally intensive SOTA DNN that can compute it correctly. 

We then evaluate the fitness of each individual on the test set: we measure both accuracy and MFLOPS for PERTINENCE, considering all operations, i.e., the input dispatcher (feature extraction and fully connected layer) and the subsequent opportunistic DNN execution.
Hence, we update the current Pareto front with the non-dominated individuals. The MOEA then evolves the current population by eliminating the least fit individuals (i.e., those with low fitness scores) and generating a new population through mutation and crossover of the chromosomes.
The MOEA runs for a specified number of epochs. At the end of execution, we evaluate the final population using a validation set that was not used during model training or the MOEA process. This evaluation allows us to obtain the final solutions.

\section{Experimental Evaluation}
\label{sec:eval}
We demonstrate our approach on state-of-the-art (SOTA) DNN models: we resort to CNNs trained on the CIFAR-10 and CIFAR-100 datasets~\cite{cifar_data} and to ViTs trained on the Tiny Imagenet datasets~\cite{tinyimage_data}.

We use THOP~\cite{thop}, Pytorch-OpCounter API to obtain the Multiply-Accumulate Operations (MACs), parameters and the resulting FLOPS for the DNN models.
During the exploration, we set the MOEA population size to 50. We used the Simulated Binary Crossover (SBX) with eta and crossover probability set to 20 and 0.9, respectively.
For the mutation, we used the Polynomial Mutation with the eta parameter set to 25. For each individual, the FC layer is trained for 20 epochs. We let penalties evolve in the range [0,100] with a step size in the range [0.5,1].
Finally, we let the MOEA evolve for 50 epochs.

Fig.~\ref{fig:DNN-pareto} shows the Pareto front of  DNN models in the space defined by the number of (floating point) operations and the Top-1 accuracy.

In the following graphs, we report the same graphs as in Fig.~\ref{fig:DNN-pareto} of the state-of-the-art DNNs (\greendot~green dots) and on top we add PERTINENCE solutions found during the exploration (\blackdot~black dots).
Finally, we highlight the points lying on the Pareto-front (\drawcross~red crosses).
As already mentioned, for every PERTINENCE solution found during the exploration, the reported number of operations includes all operations, i.e., the input dispatcher (feature extraction and fully connected layer) and the subsequent opportunistic DNN execution. We report the average number of operations per image across the validation set. 

All the experiments were executed on a 48-core Intel(R) Xeon(R) Gold 5220R processor running at 2.2 GHz and a NVIDIA A100 80GB PCIe GPU.
We performed extensive experiments to explore the PERTINENCE capabilities. For CNNs, first we chose a SOTA network $i$ from the Pareto-front (see Fig.~\ref{fig:DNN-pareto}) as a feature extractor and trained the fully connected layer of the input dispatcher to dispatch the inputs to different subsets of the SOTA networks having equal or better accuracy than $i$ -- for instance when using \texttt{resnet14} to extract features, we do not dispatch to \texttt{resnet8}, instead we dispatch to different subsets of more (or equally) accurate NNs, e.g., \{\texttt{shufflenetv2\_x0\_5} and \texttt{vgg16\_bn}\}, or \{\texttt{resnet14} and \texttt{vgg16\_bn}\}, etc.
Specifically, we explored over 25 subsets, resulting in more than 400 solutions. For ViTs, we used the \texttt{resnet50} CNN backbone as feature extractor, given its small relative dimension compared to ViTs, i.e., 2\% of the MFLOPs compared to the smallest ViT considered (\texttt{DeiT-S}) and 0.055\% compared to the largest (\texttt{ViT-L}). 
For space constraints, we only show a subset of the exploration results leading to solutions dominating the SOTA networks or lying on the Pareto front.

It is to be noted that the Pareto fronts shown in Figures 6 to 10 are the result of a single GA run per model combination. Within each run, the MOEA evaluates a population of individuals across 50 generations, each individual corresponding to a distinct combination of penalty values in the range [0,100] with step size [0.5,1] and a weighting scheme (INS, ISNS, or ENS). At the end of the run, all evaluated solutions are ranked and the subset of non-dominated solutions,  those not strictly outperformed by any other solution in both accuracy and MFLOPs simultaneously, are retained to form the reported Pareto front.

\input{cifar10-Resnet8-feature-extract}
\input{cifar10-Resnet14-feature-extract}
\begin{figure}[h]
    \centering
\includegraphics[width=.9\columnwidth]{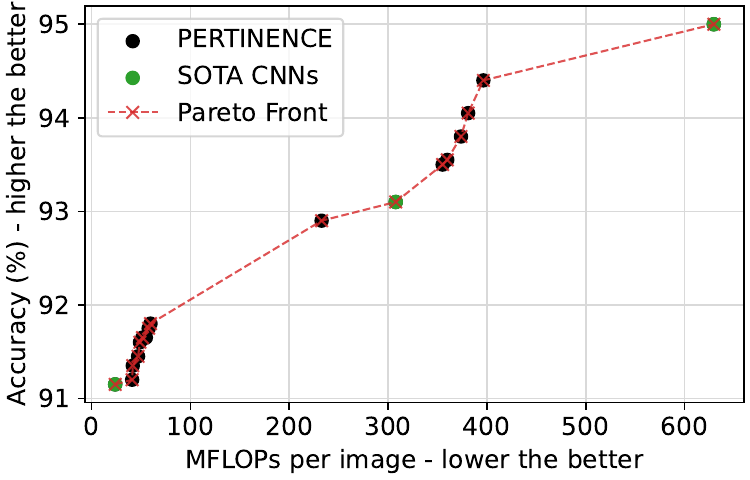}
    \caption{
    Solutions obtained using \texttt{shufflenetv2\_x0\_5} feature extraction capabilities on CIFAR-10 dataset and CNNs. Inputs are dispatched to either \texttt{shufflenetv2\_x0\_5} or \texttt{vgg16\_bn}.}
    \label{fig:cf10-res-Snetf}

\end{figure}
\input{cifar100-shufflenet-feature-extract.tex}
\input{tinyimagenet-resnet50-feature-extract.tex}
\input{power.tex}

\subsection{Results with CNNs on CIFAR-10 dataset}

\begin{table}[b]
\caption{Results obtained when using \textit{resnet8} as feature extractor and dispatching inputs to  either \textit{shufflenetv2\_x0\_5} (SN) or \textit{vgg16\_bn} (VGG16). In bold, solutions dominating SOTA CNNs.}
    \label{tab:example-exploration-resnet8extr}
    \centering
\begin{adjustbox}{width=\columnwidth}
\begin{tabular}{cccccc}
\multicolumn{3}{c|}{\multirow{2}{*}{\scriptsize \begin{tabular}[c|]{@{}c@{}}VGG16: 95\% accuracy, 1259320 MFLOPS\\VGG11: 93.10\% accuracy, 615600 MFLOPS\end{tabular}}} & \multicolumn{3}{c}{
Penalties*} \\
\multicolumn{3}{c|}{} & 10 / 0.001 & 10 / 0.005 & 10 / 0.05 \\ \hline
\multirow{6}{*}{\begin{tabular}[c]{@{}c@{}}Weighting\\ scheme\end{tabular}} & \multirow{2}{*}{ENS} & \multicolumn{1}{c|}{Accuracy} & \textbf{93.45} & 92.90 & 92.10 \\
 &  & \multicolumn{1}{c|}{MFLOPS} & \textbf{497025.9} & 314083.4 & 169304.4 \\\cline{2-6}
 & \multirow{2}{*}{INS} & \multicolumn{1}{c|}{Accuracy} & \textbf{95.20} & 92.49 & 92.50 \\
 &  & \multicolumn{1}{c|}{MFLOPS} & \textbf{802334.0} & 283795.1 & 283794.9 \\\cline{2-6}
 & \multirow{2}{*}{ISNS} & \multicolumn{1}{c|}{Accuracy} & \textbf{93.85} & \textbf{93.10} & 92.40 \\
 &  & \multicolumn{1}{c|}{MFLOPS} & \textbf{557602.9} & \textbf{382535.4} & 261381.4\\\hline
\multicolumn{6}{r}{\scriptsize*VGG predicted as SN / SN predicted as VGG}\\
\end{tabular}
\end{adjustbox}
\end{table}

We first show the results of the proposed approach on CIFAR-10 dataset with CNNs. 
In Fig.~\ref{fig:cf10-res},~\ref{fig:cf10-res-r14f}, and~\ref{fig:cf10-res-Snetf}, we report the solution space obtained using \texttt{resnet8}, \texttt{resnet14}, and \texttt{shufflenetv2\_x0\_5}, respectively, as the feature extractors.
As reported in the figures, we selected different subsets of CNN models from the Pareto-front (see Fig.~\ref{fig:cf10-pareto}) to which the input dispatcher can dynamically provide inputs.

We report cases where the input are dispatched to two CNN models (Fig.~\ref{fig:cf10-res}(a)-(c), Fig.~\ref{fig:cf10-res-r14f}(a)-(b), and Fig.~\ref{fig:cf10-res-Snetf}), to three CNN models (Fig.~\ref{fig:cf10-res}(d)-(e) and Fig.~\ref{fig:cf10-res-r14f}(c)) and four CNN models (Fig.~\ref{fig:cf10-res}(f)).
Note that in Fig.~\ref{fig:cf10-res-r14f} we do not show \texttt{resnet8} and in Fig.~\ref{fig:cf10-res-Snetf} also \texttt{resnet14}: indeed, as also mentioned earlier, extracting features with \texttt{resnet14} makes \texttt{resnet8} useless since \texttt{resnet14} is more accurate and is supposed to be able to handle all inputs \texttt{resnet8} can handle; the same applies to \texttt{shufflenetv2\_x0\_5} w.r.t. \texttt{resnet14}.

As a first observation, the exploration consistently found solutions where PERTINENCE is capable of providing a MFLOPS-accuracy trade-off lying on the Pareto-front of the explored space. This means that PERTINENCE is firstly useful for ML users to have \textit{alternative choices} to existing state-of-the-art models 
by dynamically combining them online.
Secondly, we can observe in Fig.~\ref{fig:cf10-res}(c) and Fig.~\ref{fig:cf10-res-r14f}(b) that the exploration found solutions where PERTINENCE could provide \textit{trade-offs dominating state-of-the-art CNN models}. Indeed, as reported in Tab.~\ref{tab:example-exploration-resnet8extr} and Fig.~\ref{fig:cf10-res}(c), \textit{the exploration led to a solution dominating \texttt{vgg16\_bn} and three dominating \texttt{vgg11\_bn}.}
While \texttt{vgg16\_bn} and \texttt{vgg11\_bn} obtain 95\% and 93.10\% top-1 accuracy, respectively, PERTINENCE was able to provide \textit{dominant trade-offs obtaining 95.2\%, 93.85\%, 93.45\%, and 93.10\% top-1 accuracy with 36.29\% and 9.4\%, 19.2\%, and 37.81\% fewer operations than the corresponding dominated point}, respectively.

Moreover, as reported in Fig.~\ref{fig:cf10-res-r14f}(b), the exploration found three solutions where PERTINENCE dominates \texttt{vgg11\_bn} CNN.
The \textit{dominant trade-off alternatives provided by PERTINENCE obtain 93.45\%, 93.4\%, and 93.15\% top-1 accuracy with 0.37\%, 0.67\%, and 12.77\% fewer operations} respectively.

\subsection{Results with CNNs on CIFAR-100 dataset}
\label{cifar-100-results}
In Fig.~\ref{fig:cf100-res}, we report the exploration results for the CIFAR-100 dataset with CNNs from the Pareto-front in Fig.~\ref{fig:cf100-pareto}.
We report the solution space obtained using \texttt{shufflenetv2\_x0\_5} as the feature extractor to train the FC layer of the input dispatcher and different subsets of CNN models from the Pareto-front in Fig.~\ref{fig:cf100-pareto}, to which the input dispatcher can dynamically provide inputs. We report results on cases where the input can be dispatched to two CNN models (Fig.~\ref{fig:cf100-res}(a)-(b)), and to three CNN models (Fig.~\ref{fig:cf100-res}(c)-(d)).

For this dataset, the exploration always led to solutions on the Pareto front of the explored space.
Moreover, as also depicted in Fig.~\ref{fig:cf100-res}(b), PERTINENCE was able to provide \textit{six alternatives dominating \texttt{mobileNetv2\_x1\_4}}. While \texttt{mobileNetv2\_x1\_4} reaches 76.9\% top-1 accuracy, \textit{PERTINENCE offered solutions achieving 77.3\%, 77.15\%, 77.15\%, 76.9\%, 77\%, and 76.9\% top-1 accuracy, with 9.61\%, 10.41\%, 12.71\%, 13.21\%, 14.96\%, and 15.76\% fewer operations}, respectively.

\subsection{Results with ViTs on Tiny Imagenet dataset}
In Fig.~\ref{fig:tinyimagenet-res}, we report the results of the exploration for the Tiny Imagenet dataset with ViTs.
As already mentioned, we used \texttt{resnet50} model pretrained on the Imagenet ILSVRC 2012 dataset~\cite{resnet50} as the feature extractor to train the FC layer of the input dispatcher. We report two different subsets of ViT models from the Pareto-front in Fig.~\ref{fig:tinyimagenet-vit-pareto} to which the input dispatcher can dynamically provide inputs. In Fig.~\ref{fig:tinyimagenetA} we report results when the input is dispatched to three ViTs, i.e., \texttt{DeiT-S}, \texttt{DeiT-B}, and \texttt{ViT-L}.
In Fig.~\ref{fig:tinyimagenetB} we report results when the input is dispatched to four ViTs, i.e., \texttt{DeiT-S}, \texttt{Swin-Base}, \texttt{DeiT-B-distilled}, and \texttt{ViT-L}.
Among the solutions lying on the Pareto-front of the explored space, the exploration found \textit{two solutions where PERTINENCE dominates \texttt{ViT-L}} in the first case shown in Fig.~\ref{fig:tinyimagenetA}. While \texttt{ViT-L} reaches 88.05\% top-1 accuracy, \textit{in one solution PERTINENCE obtains 88.11\% top-1 accuracy, with 3.99\% fewer operations and in the other achieves 88.05\% with 5.49\% fewer operations}.
In the case shown in Fig.~\ref{fig:tinyimagenetB}, the exploration found \textit{five solutions where PERTINENCE dominates \texttt{CaiT-S36} and one where it dominates \texttt{ViT-L}}. While \texttt{CaiT-S36} reaches 86.72\% top-1 accuracy, \textit{PERTINENCE achieves
87.33\%, 87.03\%, 86.96\%, 86.91\%, and 86.83\% top-1 accuracy, with 9.27\%, 14.97\%, 18.55\%, 25.83\%, and 28.07\% fewer operations}, respectively. Moreover, compared to \texttt{ViT-L} (88.05\% accuracy), \textit{PERTINENCE obtains 88.14\% top-1 accuracy, with 13.65\% fewer operations.}

\subsection{Power consumption measurements}
To confirm the advantages that PERTINENCE achieves, we measured the power consumption of the solutions found in some of the explorations conducted with the MOEA. To do that, we deployed PERTINENCE on a embedded GPU (Jetson ORIN AGX Xavier) platform and used the \texttt{tegrastat} utility to obtain both CPU and GPU power consumption.

In figure~\ref{fig:power}, we report the trade offs between power consumption (GPU + CPU) and accuracy for two explorations we conducted for the CIFAR-10 dataset. In particular, we measured power for the results reported in Figure~\ref{fig:cifar-10-res8-c} and Figure~\ref{fig:cf10-res-r14f}b. As shown, the measured power consumption is in line with the results obtained for MFLOPS, confirming the ability of PERTINENCE to provide solutions comparable to and even dominating SOTA DNNs.

\subsection{General discussion}
The exploration results show that PERTINENCE can provide a large variety of solutions, offering different trade-offs between accuracy and number of operations, with most of them lying on the Pareto-front of the explored space.
We noticed that the choice of the DNNs impacts the final result. For example, in Figure~\ref{fig:tinyimagenetA} we let PERTINENCE use \texttt{DeiT-S}, \texttt{DeiT-B} and \texttt{ViT-L} to process the inputs and while alternatives dominating \texttt{ViT-L} were found, \texttt{CaiT-S36} and \texttt{DeiT-B-distilled} still dominated most of PERTINENCE solutions. However, when including \texttt{DeiT-B-distilled} in the exploration, PERTINENCE could provide points dominating \texttt{CaiT-S36} and also a better point dominating \texttt{ViT-L}. The same observation can be made for results in Figure~\ref{fig:cf10-res}(a) vs~\ref{fig:cf10-res}(c). 

Thus, to achieve better results, it's important to include in PERTINENCE a variety of SOTA DNNs that span different points on the target Pareto front.
For all the experiments presented in Section~\ref{sec:eval}, the input dispatcher is invoked for every image, since each image in the evaluation dataset is independent and exhibits no similarity with its preceding or succeeding images. In contrast, in Section~\ref{sec:trafic}, we demonstrate the benefits of using PERTINENCE in a practical deployment scenario -- road occupancy estimation -- where the input frames exhibit some temporal similarity. In such cases, the input dispatcher does not need to be executed for every frame; instead, it is called only when there is a significant change in the scene content.

\subsection{Dispatcher Overhead Details}
While the dispatcher impact on both accuracy and computational complexity is taken into account in all reported results, Table~\ref{tab:dispatcher_overhead} summarizes the dispatcher overhead for all configurations. An important implementation detail further reduces the effective dispatcher overhead: when the feature extractor used in the dispatcher is itself one of the candidate models on the Pareto front (e.g., ResNet8 for CIFAR-10), and the dispatcher selects that same model as the most suitable for a given input, only the fully connected (FC) layer needs to be executed additionally, the feature extraction computation is shared. In such cases, the dispatcher overhead reduces to just the FC layer cost, which is negligible.

\begin{table}[t]
\color{black}
\centering
\caption{Dispatcher Overhead for PERTINENCE}
\label{tab:dispatcher_overhead}
\renewcommand{\arraystretch}{1.3}
\setlength{\tabcolsep}{2pt}
\begin{tabular}{llcc}
\hline

\textbf{Dataset} & \makecell{\textbf{Feature}\\\textbf{Extractor}} & \makecell{\textbf{Dispatcher}\\\textbf{Overhead}\\\textbf{Per Image}\\\textbf{(MFLOPs)}} & \makecell{\textbf{Dispatcher}\\\textbf{Overhead (MFLOPs)}\\\textbf{(when selected model = }\\\textbf{the Feature Extractor)}}\\
\hline
CIFAR-10  & ResNet8      & 5.94    & 0.04 \\
CIFAR-10  & ResNet14     & 13.50  & 0.04 \\
CIFAR-10  & ShuffleNetV2 & 24.00   & 0.10 \\
CIFAR-100 & ShuffleNetV2 & 24.17   & 0.10 \\
Tiny-ImageNet & ResNet50 & 73.54   &   NA   \\
\hline
\end{tabular}%
\end{table}

\bgroup\color{black}
\subsection{Dispatcher Robustness Analysis}
To characterize the robustness of PERTINENCE's dispatcher, we analyze two representative solutions from Figure~\ref{fig:cf10-res} (c) on CIFAR-10 using ResNet8 as the dispatcher with \texttt{vgg16\_bn} (VGG) and ShuffleNetV2~$\times$0.5 (SNetv2) as candidate inference models, evaluated on 2000 validation images. Table~\ref{tab:disp_robustness} presents the dispatcher prediction confusion matrix. 
It is important to highlight that the two PERTINENCE solutions are compared to an \textbf{ideal dispatcher}, i.e., a dispatcher incurring 0 MFLOPs cost and achieving 100\% accuracy, hence realistically impossible to obtain.
The ideal dispatcher assigns 114 images to VGG (i.e., those that can be correctly classified only by VGG) and the remaining 1886 images to SNetv2.

\begin{table}[b]
\color{black}
    \centering
    \caption{Dispatcher Prediction Confusion Matrix for two representative PERTINENCE solutions on CIFAR-10 dataset }
    \label{tab:disp_robustness}
    \setlength{\tabcolsep}{2pt}
    \resizebox{\columnwidth}{!}{%
    \begin{tabular}{l c c c c}
    \textbf{Solution} & \textbf{VGG - VGG} &       \makecell{\textbf{VGG - SNetv2}\\(Accuracy loss)} &  \makecell{\textbf{SNetv2 - VGG}\\(MFLOPs waste)} & \textbf{SNetv2 - SNetv2}\\
    \hline
    \makecell[l]{Ideal\\Dispatcher} & 114 & 0 & 0 & 1886\\\hline
    \makecell[l]{PERTINENCE\\Solution 1} & 81 & 33 & 1145 & 741\\\hline
    \makecell[l]{PERTINENCE\\Solution 2} & 54 & 60 & 768 & 1118\\
    \hline
    \end{tabular}%
    }
\end{table}

When the PERTINENCE dispatcher misdispatches the image, two distinct phenomena can occur, each with asymmetric consequences. (a) \textit{Underestimation:} routing an image for which a large model is needed to a lighter model directly reduces accuracy with respect to an ideal dispatcher; Solution 1 incurs 33 such errors (1.65\% accuracy drop w.r.t. ideal), and Solution 2 incurs 60 (3.0\% accuracy drop w.r.t. ideal). (b) \textit{Overestimation:} Routing an image for which a light model is enough to a large model wastes computation with respect to an ideal dispatcher but largely preserves accuracy. Of the 1145 overestimated images in Solution 1, 1075 remain correctly classified, and of the 768 in Solution 2, 721 are correctly classified. This asymmetry is a deliberate property of the GA-optimized penalty matrix, which explicitly encodes the relative cost of each scenario. 

To better illustrate the two phenomena, Figure~\ref{fig:disp_robust} presents the distance between the two mentioned PERTINENCE solutions and the ideal solution that would be obtained if an ideal dispatcher existed (\textit{optimal}, in the figure). We show such distance along the horizontal distance, quantifying the MFLOPs overhead incurred by overestimation errors, and the vertical distance that captures the accuracy loss caused by underestimation 
errors. Solution 1, trained with a higher underestimation penalty, lies closer to the optimal along the accuracy axis (-1.65\%) but farther along the MFLOPs axis, whereas Solution 2, trained with a higher overestimation penalty, trades an additional 1.35\% accuracy drop to recover on an average 122.4 MFLOPs per image , confirming that the 
GA-optimized penalty matrix provides direct, interpretable control over the position of each solution along the accuracy-efficiency tradeoff path. In comparison, the two SOTA networks, \texttt{vgg11\_bn} and \texttt{vgg16\_bn}, lie even farther from the optimal solution, and are dominated by PERTINENCE solutions.

The \textit{dispatcher hit rate}, i.e.,  the percentage of images for which the dispatcher selects the least computationally complex model that correctly classifies the image, is 41.1\% for Solution 1 and 58.6\% for Solution 2. Despite the modest hit rate, Solution 1 achieves 95.2\% accuracy at 36.3\% fewer MFLOPs than \texttt{vgg16\_bn} alone, demonstrating that overestimation errors are computationally costly but largely accuracy-neutral. 
In the worst case, a very poorly trained dispatcher that always mis-dispatches would systematically overestimate (i.e., select a large model when a smaller one would be enough) and underestimate (i.e., select a small model when a larger one is needed). This would lead to a worst-case scenario in which the final accuracy is comparable to the small model's accuracy, and the MFLOPs are comparable to the large model's MFLOPs. This case is clearly very extreme and was never observed in practice during our experiments.

\begin{figure}[t]
    \centering
    \includegraphics[width=0.99\linewidth]{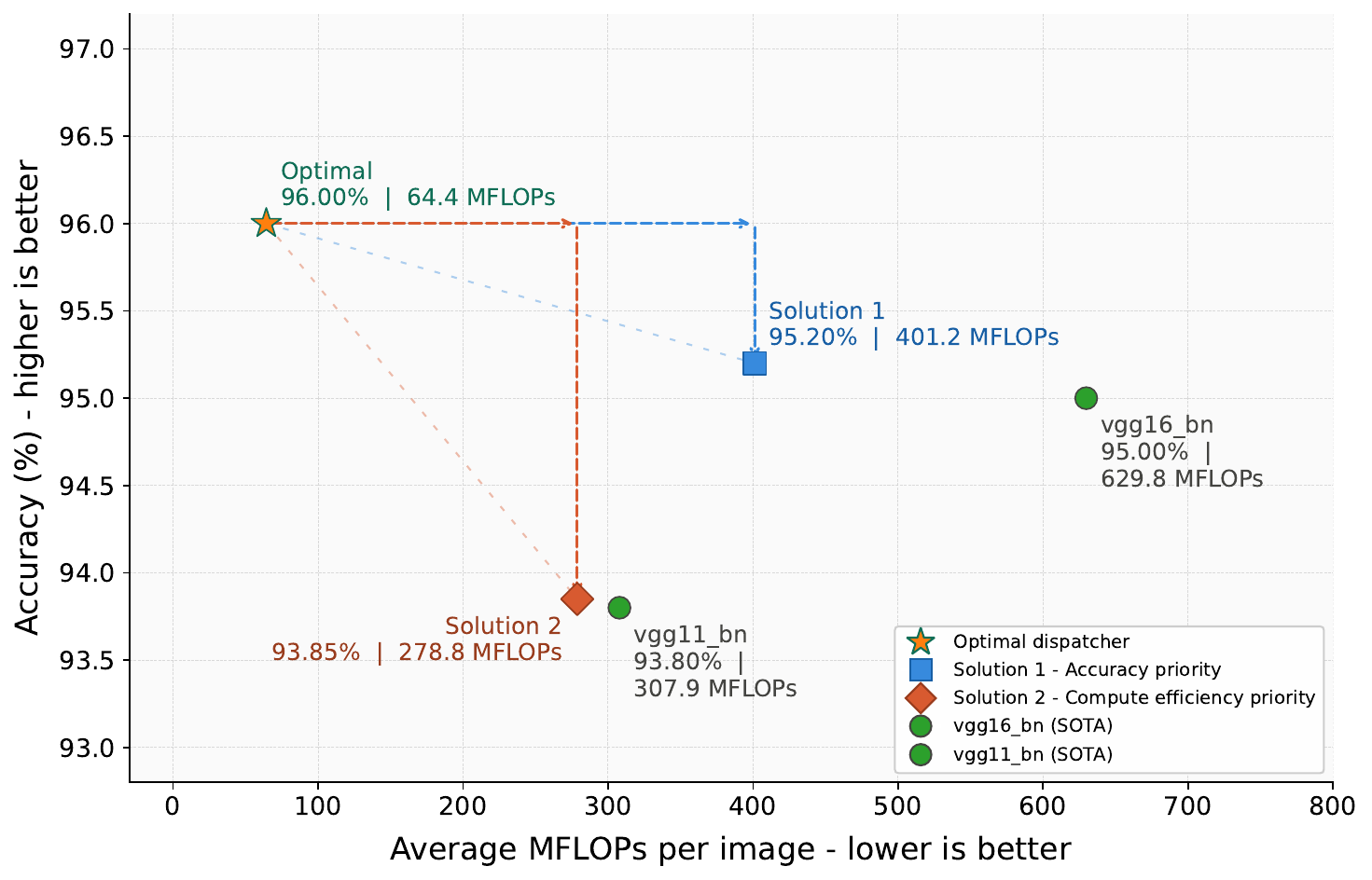}
    \caption{Accuracy vs.\ MFLOPs distance between two representative PERTINENCE solutions (from Figure~\ref{fig:cf10-res}(c)) and the optimal dispatcher on the CIFAR-10 dataset. }
    \label{fig:disp_robust}
\end{figure}

\egroup

\section{Application Case Study:  Efficient Road Occupancy Estimation Using Runtime DNN Model Selection on Traffic Camera Feeds}
\label{sec:trafic}
This section demonstrates the use of the proposed PERTINENCE approach in a real-time road occupancy estimation application that processes video feeds from cameras installed at the road's intersection.  The system leverages the city’s existing traffic surveillance infrastructure to generate vehicle count reports, categorizing detected entities by mode -- \textit{cars, buses, motorcycles, trucks, bicycles,} and \textit{pedestrians}. Based on the number and size of these detected vehicles, it estimates the proportion of the road that is occupied at any given time.  The video feeds operate at 15 frames per second (FPS), and each frame is processed by an object detection model that identifies and labels the different types of vehicles in the scene. These detections are then used to compute the road occupancy, which is further categorized into three traffic levels: \textit{low}, \textit{medium}, and \textit{high}. We use thresholds for this categorization: occupancy $\leq$15 is classified as low traffic, $>15$ and $\leq39$ as medium traffic, and $>39$ as high traffic.
For this study, we use real-time video feeds collected from traffic cameras placed at actual intersections~\cite{traffic-data}. These video streams, captured under live traffic conditions, are provided  as input to the PERTINENCE-based inference pipeline deployed on the 
Nvidia Jetson AGX Orin embedded platform. This setup faithfully emulates 
a real-time traffic monitoring deployment, where the video content 
reflects genuine real-world traffic scenarios and the inference pipeline 
processes each frame under the timing constraints imposed by the 15 FPS 
feed (66.67 ms inter-frame arrival time). The Nvidia Jetson AGX Orin 
represents the class of embedded GPU hardware typical of edge deployment 
scenarios. The power consumption of the complete pipeline is measured 
using the \texttt{tegrastats} utility, capturing both CPU and GPU power 
consumption during inference.

We used four pretrained CNN-based object detection models \texttt{TinyYOLOv3}, \texttt{YOLOv3\_320}, \texttt{YOLOv3\_416}, and \texttt{YOLOv3\_608}, trained in~\cite{traffic-data} to perform the same task, offering different accuracy scores and MFLOPS on a traffic video dataset~\cite{traffic-data}.
\begin{table}[b]
\caption{Computational cost and accuracy of the different object detection models}
    \label{tab:yolo}
    \centering
    \begin{tabular}{c|c|c}
      \textbf{Model}   & \textbf{MFLOPs/frame} & \textbf{Accuracy}\\
      \hline
      \textit{TinyYOLOv3}   &  2726 & 63.68\\
      \hline
      \textit{YOLOv3\_320}   & 19331 & 64.12 \\
      \hline
      \textit{YOLOv3\_416}   & 32670 & 73.20\\
      \hline
       \textit{YOLOv3\_608}  & 69787 & 75.41 \\
       \hline
    \end{tabular}
    
\end{table}
Table~\ref{tab:yolo} provides the computational cost in MFLOPS and inference accuracy of these models evaluated on the validation dataset consisting of 1800 frames. Table~\ref{tab:hardware_specs} provides the details of the hardware and system used for the application study experiments.

\begin{table}[t]
\centering
\color{black}
\caption{Hardware and System Specifications}
\label{tab:hardware_specs}
\renewcommand{\arraystretch}{1.3}
\begin{tabular}{ll}
\hline
\multicolumn{2}{l}{\textbf{Embedded Platform}} \\
\hline
Platform        & Nvidia Jetson AGX Orin \\
GPU             & \makecell[l]{Ampere Architecture, 1792 CUDA\\\& 56 Tensor cores} \\
CPU             & 12-core Arm Cortex-A78AE v8.2 64-bit \\
Memory          & 32 GB unified LPDDR5 memory \\
TensorRT        & 8.5.2.2\\
Power Monitor   & Tegrastats utility (CPU + GPU) \\
\hline
\multicolumn{2}{l}{\textbf{Video Feed Specifications}} \\
\hline
Frame Rate          & 15 FPS \\
Inter-frame Arrival Time  & 66.67 ms \\
Real-time Deadline  & 66.67 ms \\
\hline
\multicolumn{2}{l}{\textbf{Dispatcher Specifications}} \\
\hline
Backbone            & ResNet8 \\
Backbone Cost       & 173 MFLOPs/frame \\
Framework           & PyTorch 1.12\\
\hline
\multicolumn{2}{l}{\textbf{Training Configuration}} \\
\hline
Optimizer           & ADAM \\
Learning Rate       & $1 \times 10^{-3}$ \\
Batch Size          & 8 \\
Epochs              & 40 \\
\hline
\end{tabular}
\end{table}

\begin{figure}[b]
    \centering
    \includegraphics[trim={25pt 20pt 20pt 15pt},clip,width=\columnwidth]{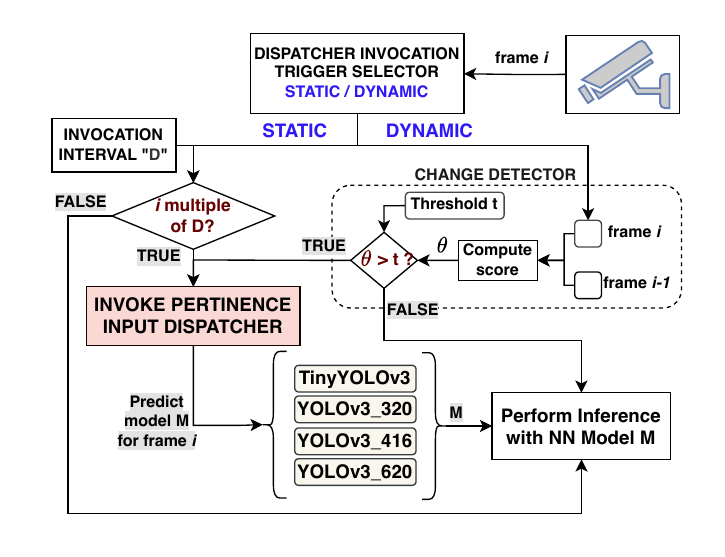}
    \caption{Road occupancy estimation NN Inference Pipeline with PERTINENCE}
    \label{fig:traffic-pipeline}
\end{figure}
Similar to the example presented in Table~\ref{tab:mot1}, out of the 1800 frames in the validation video set, the smallest model,  \textit{TinyYOLOv3}, correctly classifies \textit{1146} frames and only \textit{57} frames require the use of the larger \textit{YOLOv3\_608} model. By using PERTINENCE, we can reduce the use of the large model when it is not necessary.
In this case study, we used a \texttt{ResNet8} model as a feature extractor in PERTINENCE's input dispatcher; we trained it from scratch along with the fully-connected layer using the \textit{ADAM optimizer with a learning rate of $1*10^{-3}$, a batch size of 8, and for a total of 40 epochs}. The computational cost of the \texttt{Resnet8} backbone is \textit{173 MFLOPs} per image, which is significantly lower than that of the object detection models used for inference. 

In this case study, we explore the opportunities offered by real-time applications to reduce the number of invocations to the PERTINENCE dispatcher. Figure~\ref{fig:traffic-pipeline} illustrates the workflow of the proposed road occupancy estimation neural network inference pipeline that integrates the PERTINENCE approach for adaptive inference. The PERTINENCE dispatcher is  invoked based on dynamic scene-change characteristics or user-defined static policies. In detail, a continuous sequence of incoming image frames (I) is provided as input to the system. Upon receiving the initial frame ($I_{1}$), the framework directly invokes the Pertinence Input Dispatcher to determine the appropriate NN model for performing the inference. This ensures the system is initialized with a suitable NN model at the start of operation. For subsequent frames ($I_{2}, I_{3}, ...$), the dispatcher invocation is governed by a Dispatcher Invocation Trigger Selector, which can operate in one of two modes:
\textit{Static} mode: The dispatcher is triggered periodically at fixed intervals defined by a user-set Dispatcher Invocation Interval (\texttt{D}). For instance, when \texttt{D = 3}, every 3 frames the dispatcher is invoked, a YOLO model is selected and used for the three frames. \textit{Dynamic} mode: The dispatcher is activated based on changes in the visual content detected in consecutive frames. This approach helps to avoid unnecessary re-dispatching when the scene remains largely unchanged. 
In the Dynamic mode, a lightweight \textit{Change Detector} continuously monitors visual differences between consecutive frames. It computes a composite score ($\theta$) based on lightweight image metrics, such as pixel and histogram differences, and, if the score exceeds a predefined threshold, re-invokes the dispatcher to predict the next inference model. This mechanism ensures that model updates occur only when necessary.
Further details related to the change detection mechanism and dynamic dispatching strategy are provided in Section~\ref{sec:stat_dyn_selection}

\begin{figure}[t]
    \centering
    \includegraphics[width=\columnwidth]{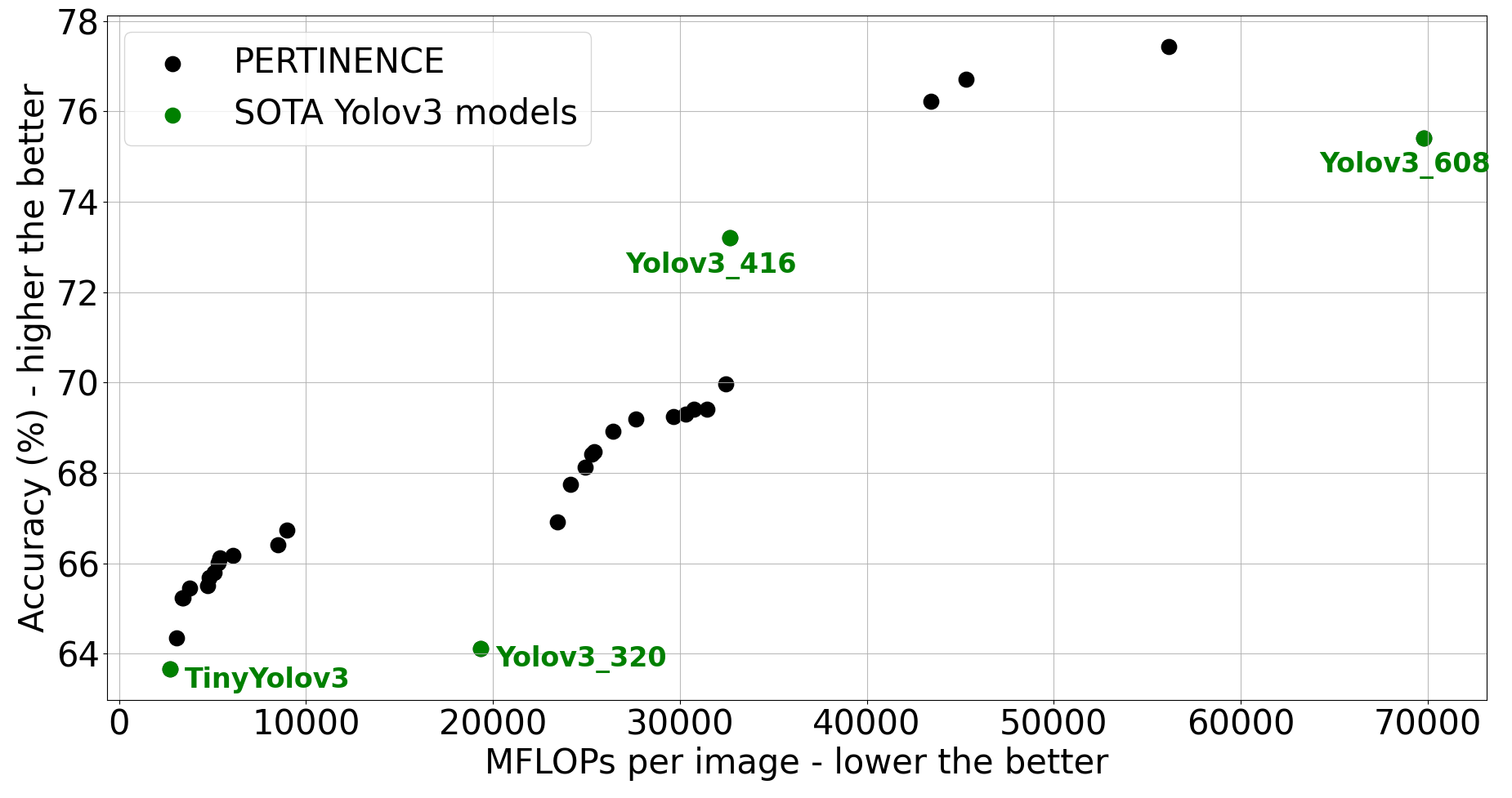}
    \caption{Solutions obtained using \texttt{resnet8} backbone as feature extractor and inputs dispatched to the appropriate object detection model by PERTINENCE}
    \label{fig:d1_sol}
\end{figure}
As with earlier experiments, we first present in Figure~\ref{fig:d1_sol} results where the input dispatcher is invoked for every image (\texttt{D=1}). 
It can be seen that PERTINENCE provides better-quality solutions, achieving higher accuracy (\textit{77.44\%, 76.71\% and 76.21\%}) than the best YOLO model (i.e., \textit{YOLOv3\_608} achieving \textit{75.41\%}), while consuming (\textit{19.5\%, 34.83\% and 37.51\%}) fewer MFLOPs. Another important observation is that  PERTINENCE achieves \textit{higher accuracy} \textit{(1.9-2.6\%)} and substantially \textit{lower MFLOPs} \textit{(72.7-53.5\%)} compared to the \textit{YOLOv3\_320} model.

\subsection{Varying input dispatcher invocation interval}

\begin{figure}[t]
    \centering
    \includegraphics[width=\columnwidth]{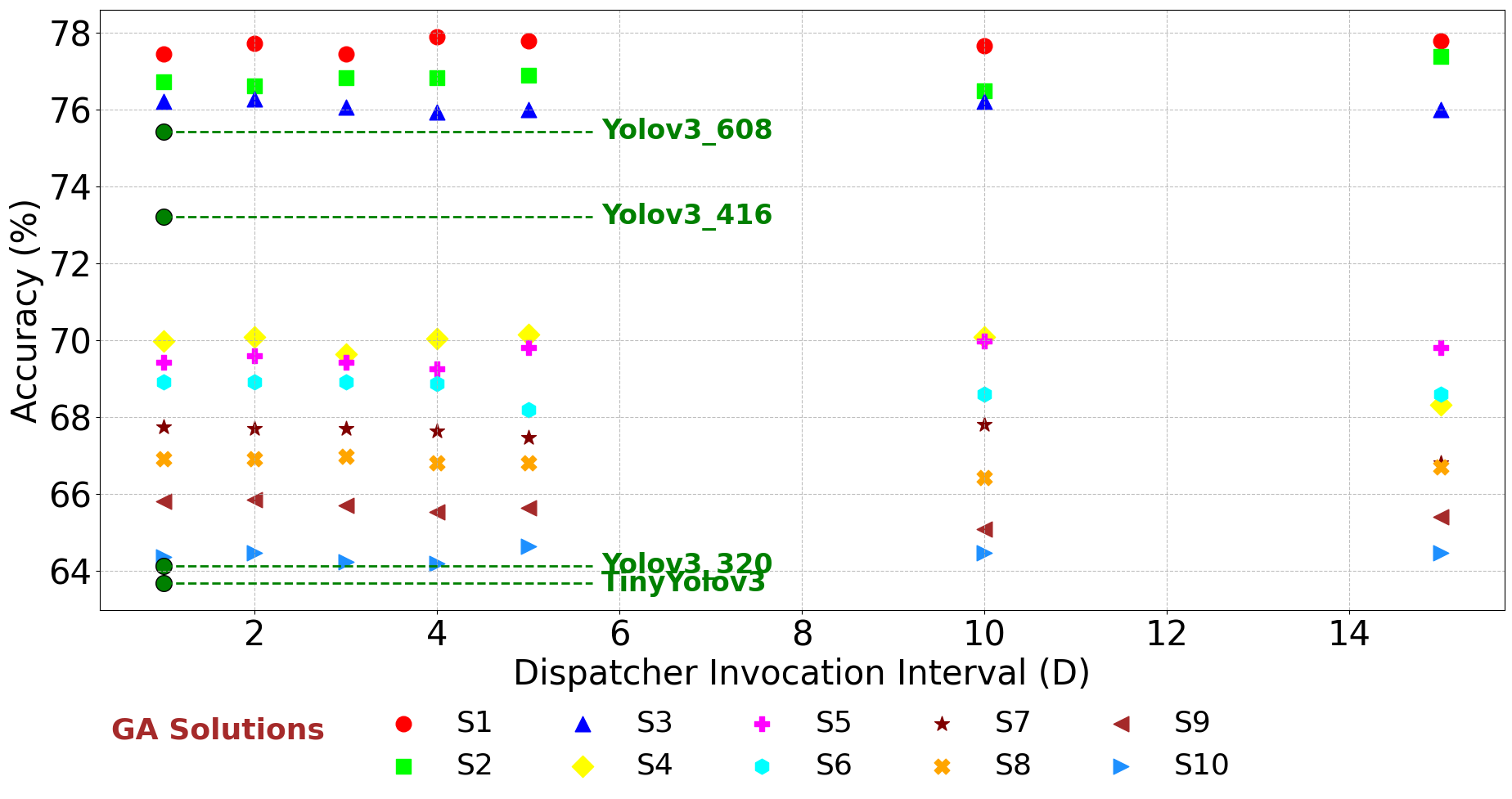}
    \caption{Tradeoffs between accuracy and input dispatcher interval (D), in terms of frames, i.e., the input dispatcher is invoked every \texttt{D} frames}
    \label{fig:d_vs_acc}
\end{figure}

\begin{figure}[b]
    \centering
    \includegraphics[width=\columnwidth]{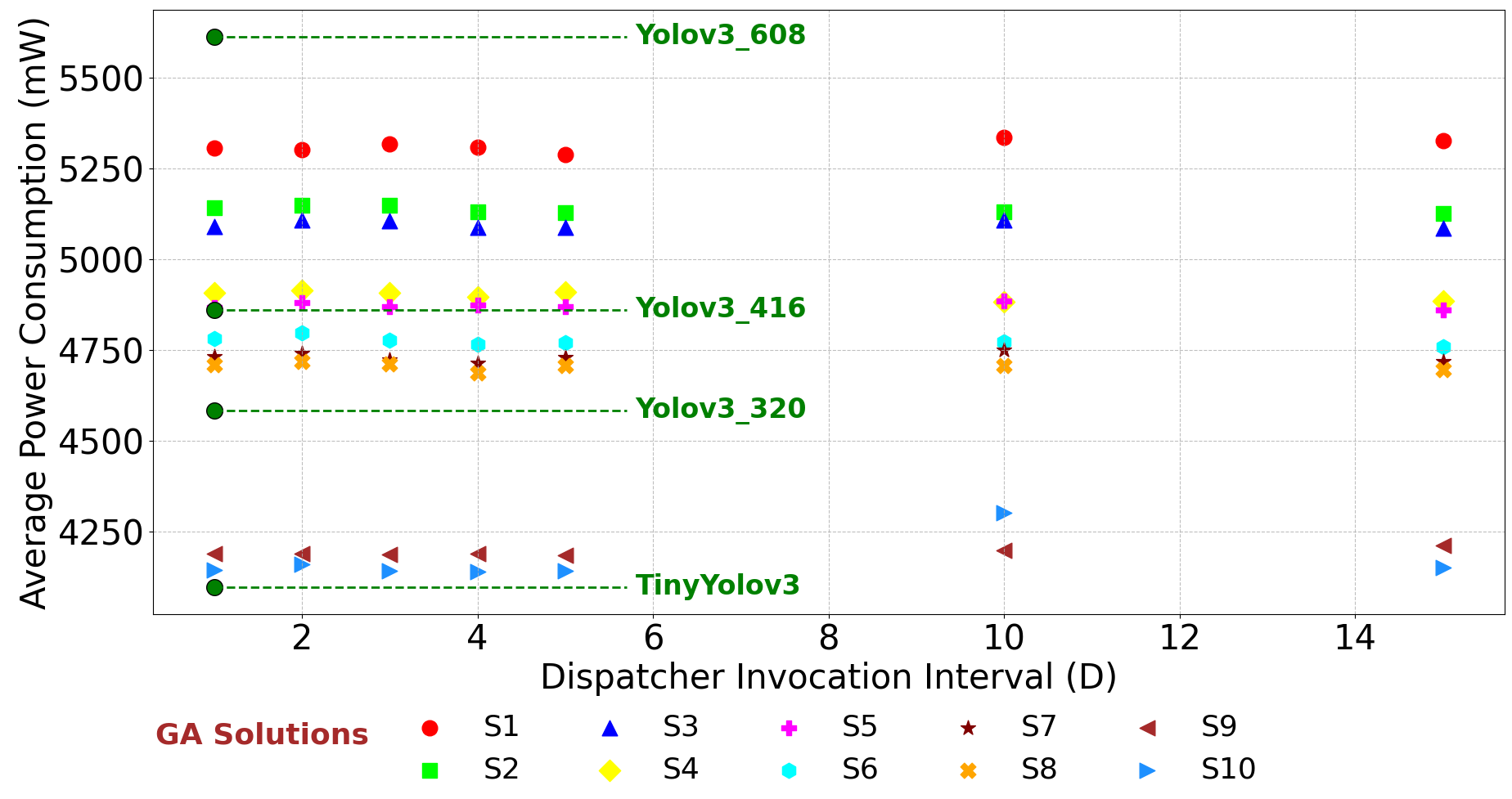}
    \caption{Variation in the Average Power Consumption(mW) with input dispatcher interval (D), in terms of frames}
    \label{fig:d_vs_power}
\end{figure}

We now analyze how the inference pipeline behaves as the interval between dispatcher invocations is varied. Specifically, we first consider a static dispatching policy, where the dispatcher is invoked once every \texttt{"D"} frames, thereby reducing the dispatching overhead while maintaining periodic control.
For this analysis, we consider a subset of 10 solutions from Figure~\ref{fig:d1_sol} generated by the genetic algorithm (GA), with each color/shape representing a distinct solution. To ensure clarity, we have chosen to limit the number of solutions shown in the graphs (Figures~\ref{fig:d_vs_acc},~\ref{fig:d_vs_power}, and~\ref{fig:d_vs_inference_time}), as including all possible solutions would reduce the overall readability. However, it is important to note that the remaining solutions exhibit the same behavior.

Figure~\ref{fig:d_vs_acc} shows the variation in observed accuracy on the validation input frames as the input dispatch interval \texttt{D} is varied. 
It can be consistently observed that the accuracy values are only slightly affected by changes in the \texttt{D} value.

\begin{figure}[t]
    \centering
    \includegraphics[width=\columnwidth]{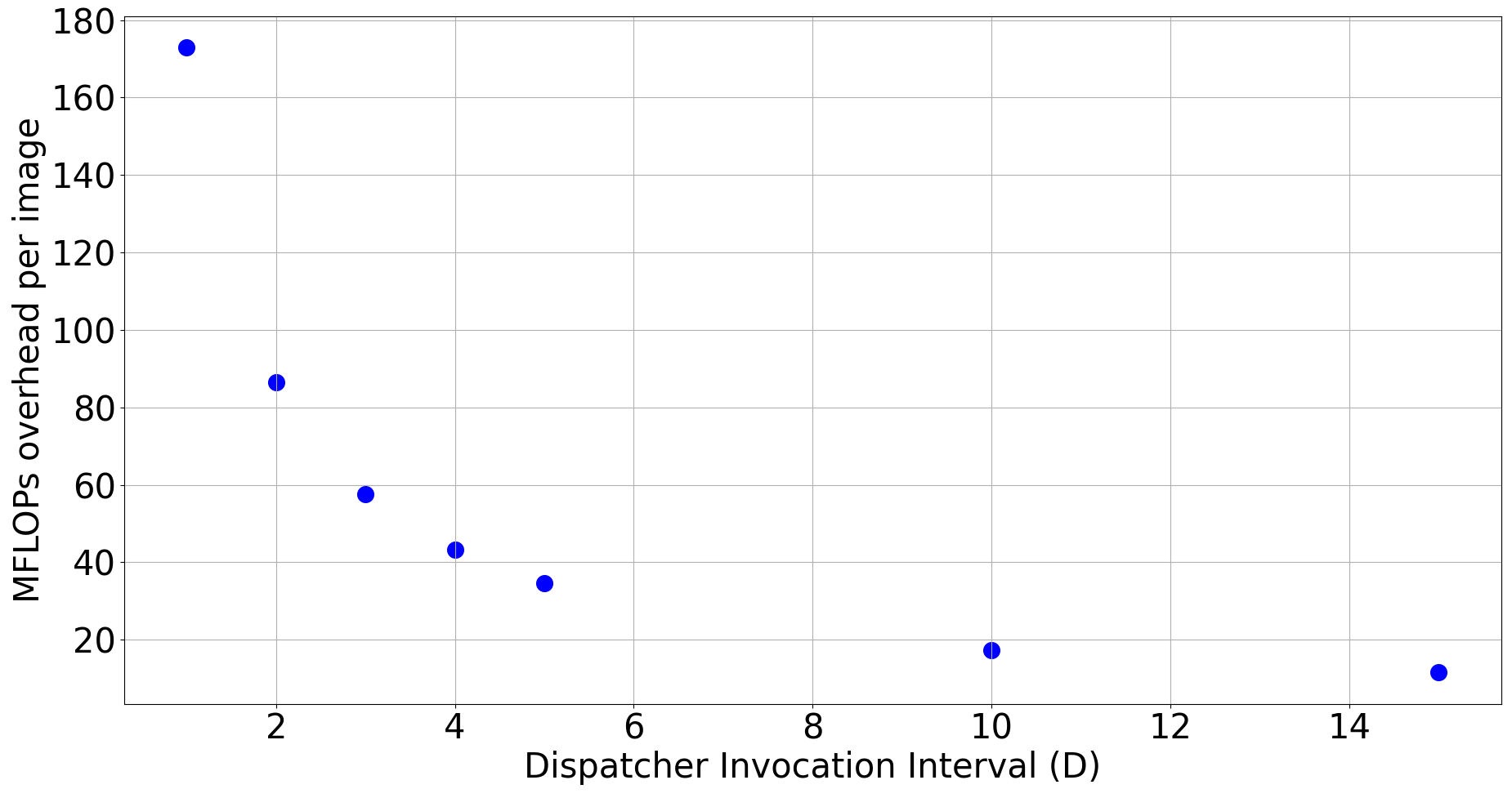}
    \caption{Variation of MFLOPs overhead per image with input dispatcher interval (D)}
 \label{fig:disp_overhead}
\end{figure}
\begin{figure}[b]
    \centering \includegraphics[width=\columnwidth]{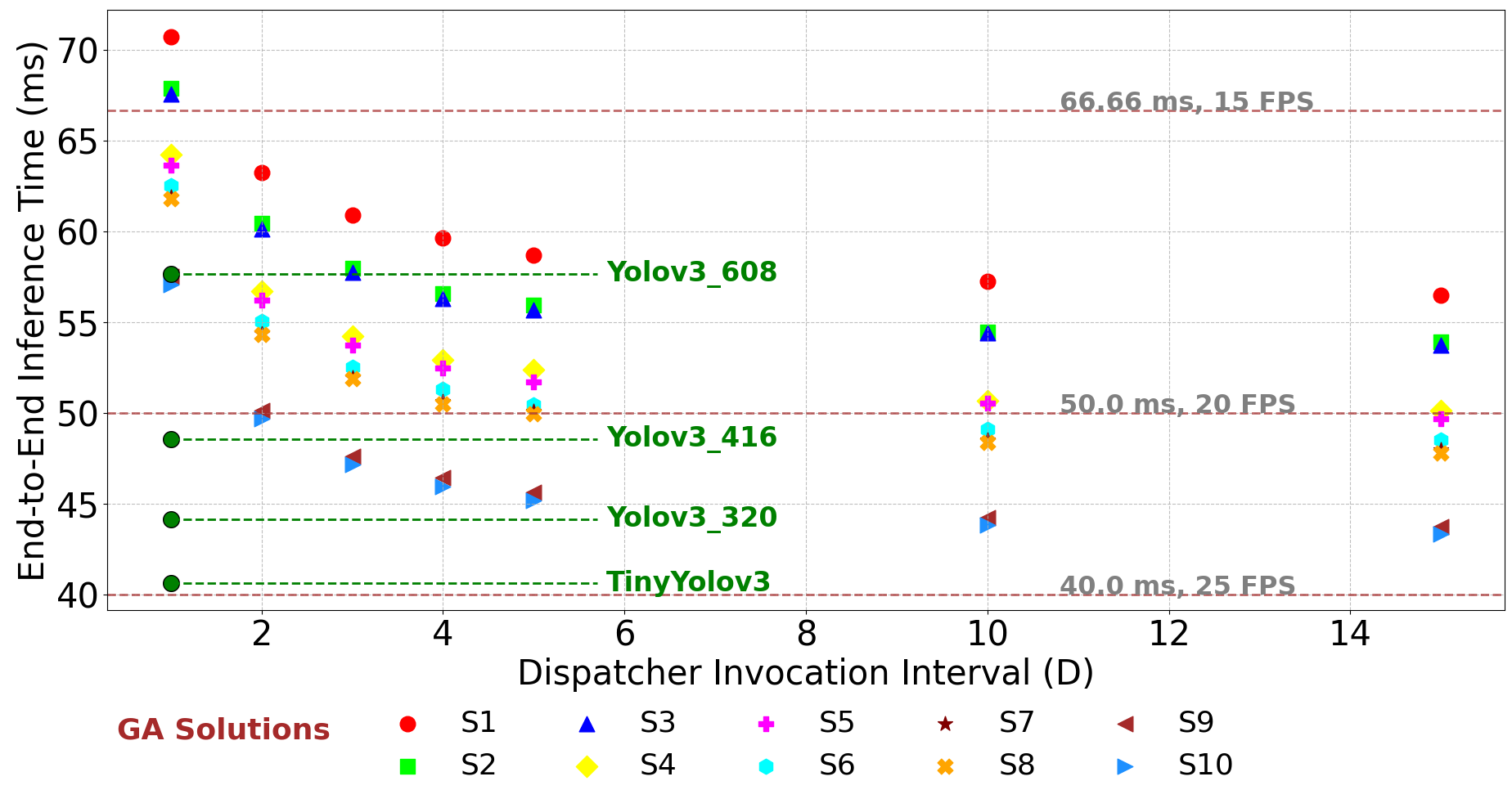} \caption{Variation in the Average Inference Time per frame(ms) with input dispatcher interval (\texttt{D}), in terms of frames}  \label{fig:d_vs_inference_time}
\end{figure}
\begin{figure}[t]
    \centering
    \includegraphics[trim={15pt 15pt 20pt 20pt},clip,width=\columnwidth]{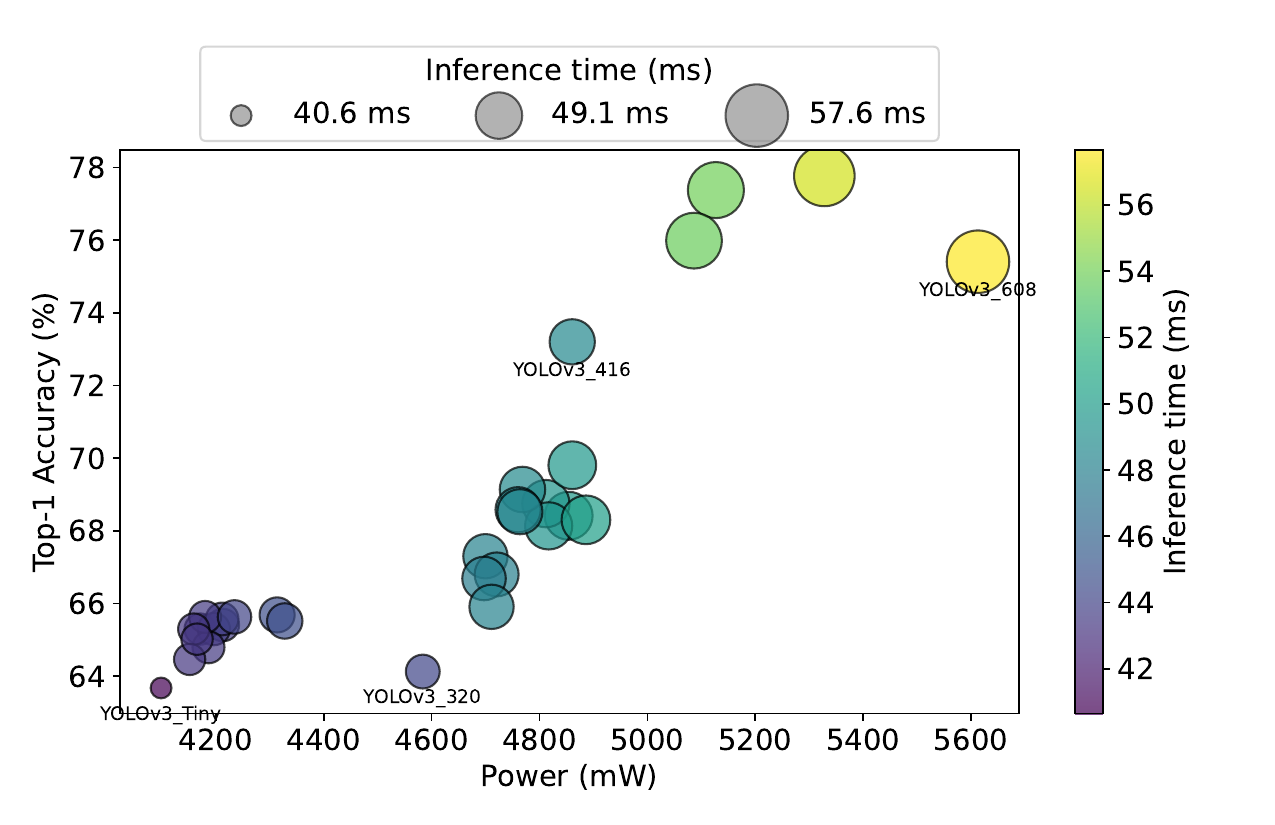}
    \caption{Trade-off among power, accuracy, and inference time obtained using PERTINENCE and dispatcher interval \texttt{D = 15}}
    \label{fig:d15}
\end{figure}

Unlike datasets such as CIFAR-10, CIFAR-100, and Tiny-ImageNet, the road traffic occupancy task requires real-time inference, requiring predictions for each camera frame. This introduces specific timing constraints, making inference latency a crucial metric for determining whether the system can meet its operational deadlines. Therefore, in addition to evaluating accuracy, we also analyze inference time to assess the system's real-time responsiveness. Furthermore, we also analyze power consumption and how varying the input dispatcher invocation interval \texttt{D} affects both the average power consumption and the average inference time.
For this analysis, we implemented the complete road occupancy estimation system (as depicted in Figure~\ref{fig:traffic-pipeline}) on the \textit{Nvidia Jetson AGX Orin} embedded development kit and used the \texttt{tegrastats} utility to measure the SoC power consumption. As mentioned earlier, the input frames arrive at a rate of 15 frames per second, corresponding to an inter-arrival time of 66.67 ms.  
As shown in Figure~\ref{fig:d_vs_power} and similar to the accuracy, the power consumption is not very sensitive to changes in the input dispatcher interval \texttt{D}, as the dispatcher's size is relatively small compared to the YOLO models.
Concerning time overhead, as shown in Figure~\ref{fig:disp_overhead}, the MFLOPs overhead per image for the input dispatcher at a specific value of \texttt{D} is $\frac{1}{D}$ of the overhead when \texttt{D} equals 1. Hence, increasing \texttt{D} results in faster inference time, as shown below. 
As already mentioned, in this case study, the MFLOPs per image for the ResNet8-based dispatcher is 173, i.e., when \texttt{D=1}, the overhead is 173 MFLOPs. When \texttt{D=15}, the MFLOPs overhead per image is $\frac{173}{15} = 11.53$. We measured its delay to be 14.82 ms; hence, $\frac{14.82}{15} = 0.99$ ms when \texttt{D=15}.
More in detail, as reported in Figure~\ref{fig:d_vs_inference_time}, for small interval values \texttt{D}, i.e., when the dispatcher is used frequently, the average inference time exceeds the real-time threshold (66.67 ms). However, since the overall system accuracy remains stable with higher \texttt{D} values (see Figure~\ref{fig:d_vs_acc}), increasing \texttt{D} can help reduce inference time, as shown in Figure~\ref{fig:d_vs_inference_time}. Moreover, for higher \texttt{D} values, the solutions obtained using PERTINENCE achieve average inference times comparable to, or even lower than, those of the best YOLO model (\texttt{YOLOv3\_608}), while simultaneously delivering higher accuracy and consuming lower power.
For example, in Figure~\ref{fig:d15}, we report -- for all the GA solutions of Figure~\ref{fig:d1_sol} -- the trade-offs among overall accuracy, power consumption, and inference time obtained for \texttt{D = 15}: \textbf{(i)} the overall accuracy of the three best solutions obtained using PERTINENCE (top-right part of the figure) is respectively \textit{2.36\%, 1.97\%, 0.58\%}  higher than \texttt{YOLOv3\_608} (which achieves 75.41\% top-1 accuracy), \textbf{(ii)} the power consumption is respectively \textit{5.07\%, 8.66\%, and 9.38\%} lower than \texttt{YOLOv3\_608} (which consumes \textit{5612.93 mW}), and \textbf{(iii)} the inference time per image is respectively \textit{2.01\%, 6.44\%, and 6.74\%} lower than \texttt{YOLOv3\_608} (which takes \textit{57.65 ms}). Moreover, compared to \texttt{YOLOv3\_320} (\textit{64.12\%} accuracy, \textit{44.14ms} inference time, \textit{4583.62mW} power consumption), PERTINENCE produces solutions achieving \textit{1.14\%} higher accuracy on average (\textit{min 0.33\%}, \textit{max 1.50\%}), while reducing average inference time by \textit{1.25\%} (min \textit{0.24\%}, max \textit{1.85\%}) and average power consumption by \textit{8.60\%} (min \textit{7.62\%}, max \textit{9.43\%}).

\subsection{Dynamic input dispatcher invocation}
\label{sec:stat_dyn_selection}
Finally, we demonstrate a dynamic scene–content–aware dispatching invocation strategy, in which the dispatcher is triggered only when a significant change in visual content is detected between consecutive frames. To achieve this, we implemented a lightweight change detector that combines two complementary metrics: \textit{Absolute Pixel Difference} and \textit{Histogram Difference}. The pixel difference metric quantifies the raw brightness variation at the pixel level, while the histogram difference metric compares the overall color and brightness distributions, enabling robust detection of scene changes even under slight camera movements.

A \textit{composite change score $\theta$} is computed as a weighted sum of these two metrics, and the dispatcher is triggered when this composite score exceeds a given threshold.
To find satisfactory values for the weights and threshold, we automatically explored more than \textit{500 different weight–threshold combinations} in a grid search fashion, each requiring approximately \textit{1 minute} to evaluate. We then chose \textit{weights of 0.2 for the pixel difference and 0.8 for the histogram difference}, along with a threshold of \textit{0.016}, after which the input is processed by the dispatcher to determine the next YOLO model to use. The selected threshold exhibits an asymmetric sensitivity profile, performance degrades gracefully for thresholds below 0.016, with accuracy varying by less than 0.7 percentage points ($pp$) across the range [0.001, 0.016], whereas exceeding this value triggers a sharp decline of approximately 2.4 $pp$ at 0.019 alone, as lower thresholds incur excessive dispatching overhead from frequent resolution switches while higher thresholds suppress adaptive switching and force over-reliance on heavier models, confirming 0.016 as the most attractive operating point that jointly minimizes per-image processing time and maximizes accuracy. Table~\ref{tab:threshold_grid_search} presents the grid search results across candidate thresholds, demonstrating that 0.016 achieves the optimal balance between accuracy, dispatching overhead, and model utilization efficiency, with performance degrading gracefully below this value and declining sharply beyond it.
\begin{table}[t]
\color{black}
\centering
\caption{Grid Search Results for Threshold Selection}
\label{tab:threshold_grid_search}
\renewcommand{\arraystretch}{1.2}
\begin{tabular}{ccc c}
\hline
\textbf{Threshold} & \textbf{Accuracy (\%)} & \textbf{Per-Image Time (ms)} & \textbf{\# Frame Changes} \\
\hline
0.001 & 77.88 & 67.78 & 1560 \\
0.004 & 77.88 & 66.81 & 1439 \\
0.007 & 78.33 & 63.28 & 995  \\
0.010 & 78.55 & 60.39 & 639  \\
0.013 & 77.83 & 57.84 & 387  \\
\textbf{0.016} & \textbf{78.27} & \textbf{56.07} & \textbf{210} \\
0.019 & 75.82 & 56.99 & 120  \\
0.022 & 75.04 & 57.85 & 60   \\
0.025 & 75.65 & 58.14 & 25   \\
\hline
\end{tabular}
\end{table}
The change detection step is extremely lightweight in terms of latency, with an average computation time of 1.349 ms per frame, ensuring that it does not add significant latency to the overall inference pipeline.

\begin{figure}[b]
    \centering
    \includegraphics[width=\columnwidth]{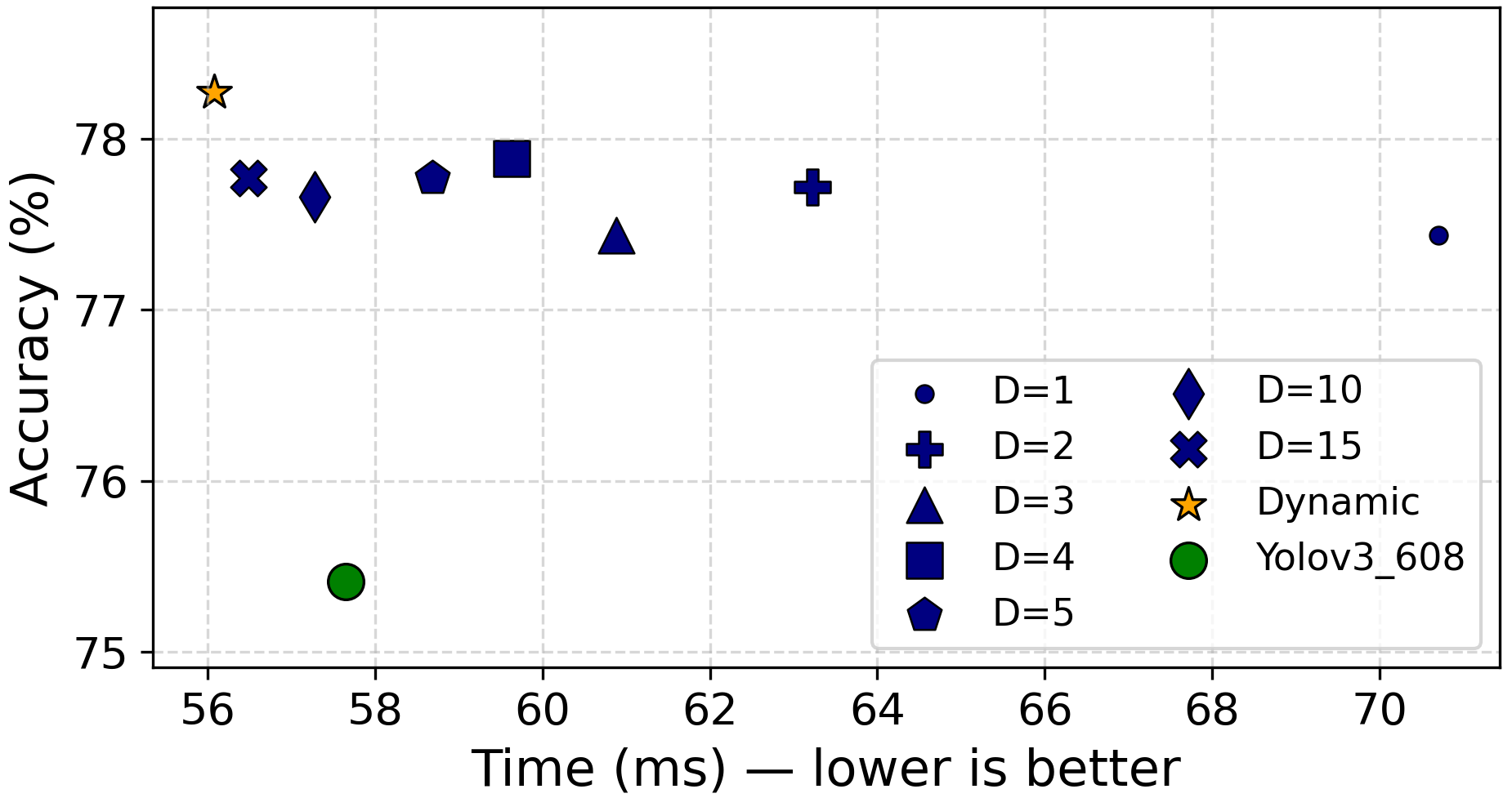}
    \caption{Variation in the Average Inference Time per frame(ms) with input dispatcher interval (D).}
    \label{fig:stat_dyn_time}
\end{figure}

\begin{figure}[t]
    \centering
    \includegraphics[width=\columnwidth]{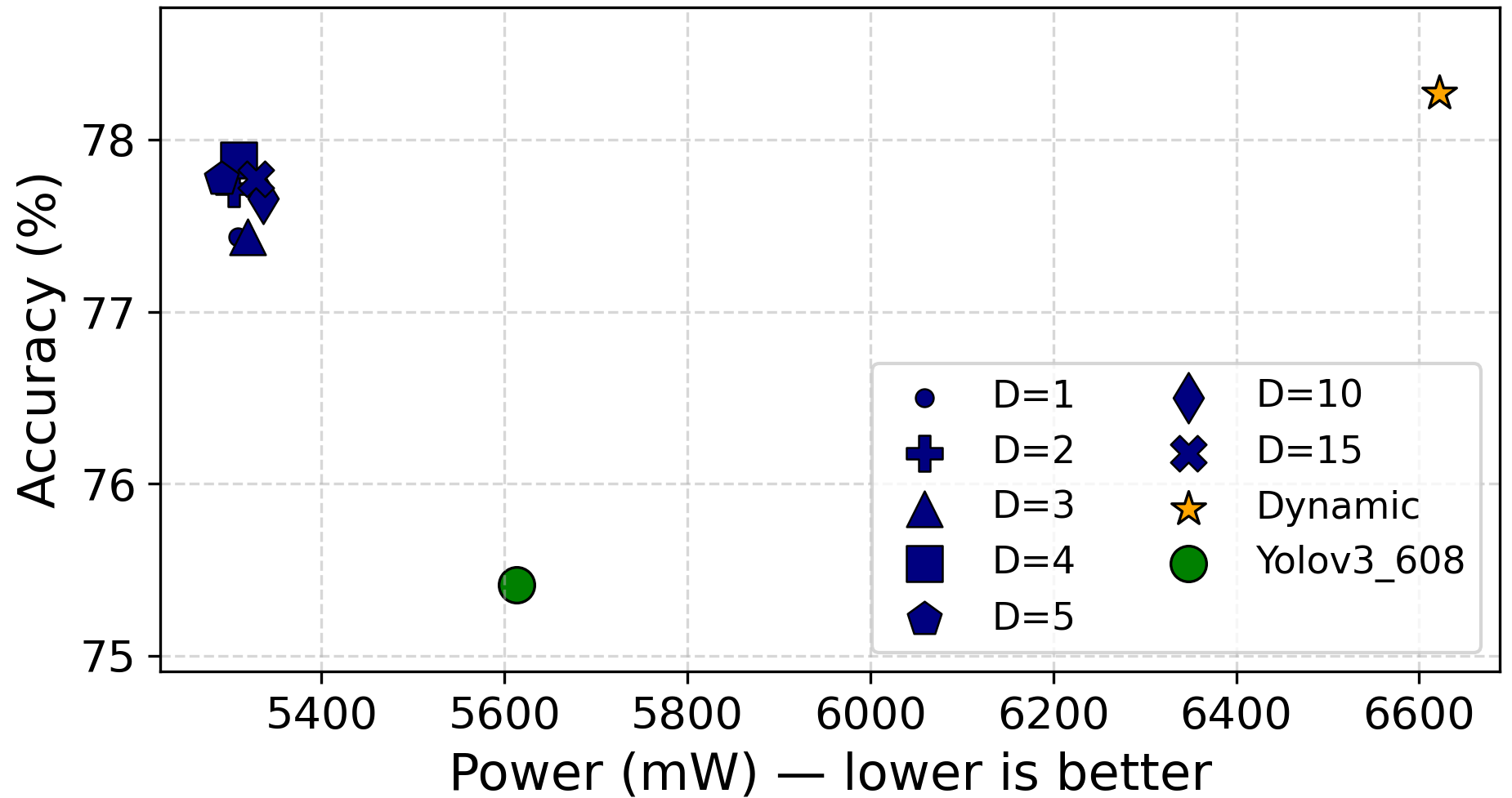}
    \caption{Variation in the Average Power Consumption(mW) with input dispatcher interval (D).}
    \label{fig:stat_dyn_power}
\end{figure}

Figures~\ref{fig:stat_dyn_time} and~\ref{fig:stat_dyn_power} report the results for the \textit{dynamic scene–content–aware} dispatching invocation compared to the static dispatcher invocation interval, both applied to the \textit{best solution}, in terms of accuracy, found through PERTINENCE, i.e., the topmost black dot~\blackdot~in Figure~\ref{fig:d1_sol}, also plotted as red dots~\reddot~in Figures~\ref{fig:d_vs_acc},~\ref{fig:d_vs_power}, and~\ref{fig:d_vs_inference_time} for differend \texttt{D} values.
The dynamic scene–content–aware dispatching invocation strategy (\yellowstar~in the graphs) achieves the highest accuracy (78.27\%) and the fastest inference time (56.07 ms), albeit with a moderate power overhead of 15\% compared to the SOTA YOLOv3\_608 model and 19.5\% compared to the static dispatcher (\texttt{D=15}). This is due to the computation of the composite change score at every frame. In contrast, the static dispatching strategy provides higher power efficiency while maintaining accuracy improvements over the SOTA model.

Overall, the static and dynamic approaches are complementary: the dynamic scheme is preferable when accuracy and latency are the primary constraints, whereas the static scheme is better suited for power-efficient deployments. In either case, both variants of the proposed PERTINENCE approach consistently outperform the SOTA baseline in terms of inference accuracy, demonstrating its applicability across diverse deployment scenarios.

\bgroup\color{black} 
\subsection{Memory Footprint Analysis}
Table~\ref{tab:memory_footprint} presents the memory footprint analysis of  PERTINENCE compared to the single best SOTA model for each dataset in this work. The incremental memory overhead is computed as the difference in GPU memory after inference between PERTINENCE and the single best SOTA model. 
\begin{table}[b]
\centering
\caption{Memory Footprint Analysis: PERTINENCE vs. Best Single Model}
\label{tab:memory_footprint}
\renewcommand{\arraystretch}{1.5}
\setlength{\tabcolsep}{2pt}
\begin{threeparttable}
\resizebox{\columnwidth}{!}{%
\begin{tabular}{l c c c c r}
\hline
\textbf{Dataset} & \textbf{Best} & \textbf{Single} & \textbf{PERTINENCE} & \textbf{PERTINENCE} & \textbf{Overhead} \\
 & \textbf{Model} & \textbf{(MB)} & \makecell{\textbf{models} \\\textbf{(incl. dispatcher)}} & \textbf{(MB)} & \textbf{(MB)} \\
\hline
CIFAR-10 
  & vgg16\_bn 
  & 1552.56 
  & \makecell{ResNet8 + \\ ShuffleNetV2$\times$0.5 +\\ vgg16\_bn }
  & 1657.54 
  & 104.98 \\
\hline
CIFAR-100 
  & repvgg\_a2 
  & 1717.21 
  & \makecell{ShuffleNetV2$\times$0.5 + \\MobileNetV2$\times$1.4 + \\repvgg\_a2}
  & 1743.07
  & 25.86 \\
\hline
Tiny-ImageNet 
  & ViT-L 
  & 3742.64 
  & \makecell{ResNet50 + DeiT-S +\\ DeiT-B + ViT-L}
  & 4542.47 
  & 799.83 \\
\hline
Traffic 
  & YOLOv3-608 
  & 1567.60 
  & \makecell{ResNet8 + \\Tiny-YOLOv3+ \\YOLOv3-320 + \\YOLOv3-416 +\\ YOLOv3-608} 
  & 4182.02 
  & 2614.42 \\
\hline
\end{tabular}%
}
\end{threeparttable}
\end{table}
\egroup
An important observation is that deep learning inference frameworks such as TensorRT allocate a significant fixed memory workspace irrespective of model size, which substantially reduces the effective incremental overhead of loading additional lightweight models. This explains why, in the case of CIFAR-100, PERTINENCE models actually consumes only 25.86 MB memory more than the best single model. 
For CIFAR-100 and CIFAR-10, the incremental overhead remains modest at 25.86-104.98 MB, a marginal cost relative to the baseline. For cases such as Tiny ImageNet and the Traffic study, the memory footprint is necessarily larger due to the inclusion of high-capacity models required to handle scene complexity, and we acknowledge this as an explicit trade-off. Critically, however, this additional memory cost is justified by measurable accuracy gains that no single model achieves independently, for the Traffic dataset in particular, PERTINENCE delivers superior accuracy precisely because it retains access to YOLOv3-608 for complex scenes while delegating simpler frames to lightweight alternatives, a capability that a single-model deployment cannot replicate. 

\section{Related Work}
Enabling deep learning models on resource-constrained devices has garnered significant attention in recent years. Techniques like pruning, quantization, knowledge distillation ~\cite{model_compression,pruning}  discussed earlier in section~\ref{sec:intro}, provide single model solutions for enabling efficient inference. However, using a fixed model across all the inputs is sub-optimal~\cite{single_subopt1}, as showcased in Table~\ref{tab:mot1}. Towards runtime solutions, some of the existing works~\cite{run-adapt1,bldnn,run-adaptmul} focus on runtime model selection. 
For example, BL-DNN~\cite{bldnn} employs a cascaded inference strategy where a little DNN performs the inference first, and a score margin-based checker determines if its output is accurate. If not, a big DNN is invoked. While this reduces computation for confident predictions with a little DNN, it incurs a higher penalty in terms of increased computations and execution time for misclassified inputs due to sequential execution through the little DNN, success checker, and then the big DNN. Similarly, the work in~\cite{run-adaptmul} uses a neural multiplexer to decide, based on the input complexity, whether to process the input locally with a smaller model or offload it to the cloud for processing with a larger model. 
SECS~\cite{run-adapt2} proposes runtime pruning of the model based on the class skew detected in the input and resource constraints. 
Finally, arecent approach~\cite{ISCAS25} proposes runtime quantization of DNNs based on image quality metrics and human-derived difficulty scores.  Cai et al.~\cite{ofa} propose the Once-for-All (OFA) network, which  trains a single supernet supporting diverse architectural settings  (depth, width, kernel size, and resolution) by decoupling training and deployment-time specialization via a progressive shrinking  algorithm. Given a target hardware platform and resource constraint, a specialized sub-network is extracted from the OFA supernet without 
additional training, enabling efficient deployment across diverse edge devices. However, OFA specialization is performed \textit{once at  deployment time} for a fixed hardware-resource scenario, the  extracted sub-network then operates as a static model during inference, 
applying uniform computation to all inputs regardless of their  complexity. In contrast, PERTINENCE dynamically selects among fully independent pre-trained models at runtime based on 
each individual input, requiring no supernet training, no architectural  coupling between candidate models, and no deployment-time specialization step.

\subsection{Comparison with Prior Work}
Among the cited studies, the most similar to PERTINENCE is the one proposed in~\cite{run-adapt1}, which employs a premodel composed of sequential k-NN classifiers to dynamically select the most suitable pre-trained CNN for each input image. This selection relies on handcrafted image features, including the \textit{number of keypoints, brightness, contrast, edge length, hue, area-to-perimeter ratio, and aspect ratio}. 
The first k-NN model (KNN-1) decides if the input image should be processed by the first DNN model (Model-1). If yes, it returns that label and the process stops. If not, the features are passed to the next k-NN model (KNN-2), which decides for Model-2, and so on.
For a fair comparison, we reimplemented this approach and evaluated it using the pre-trained models on the CIFAR-100 dataset (\texttt{mobilenetv2\_x0\_75}, \texttt{mobilenetv2\_x1\_4}, \texttt{shufflenetv2\_x0\_5} and \texttt{repvgg\_a2}) as presented in Section~\ref{cifar-100-results} (Figures~\ref{fig:cf100-res}(a)-(d)). 

\begin{figure}[b]
    \centering
    \includegraphics[width=\columnwidth]{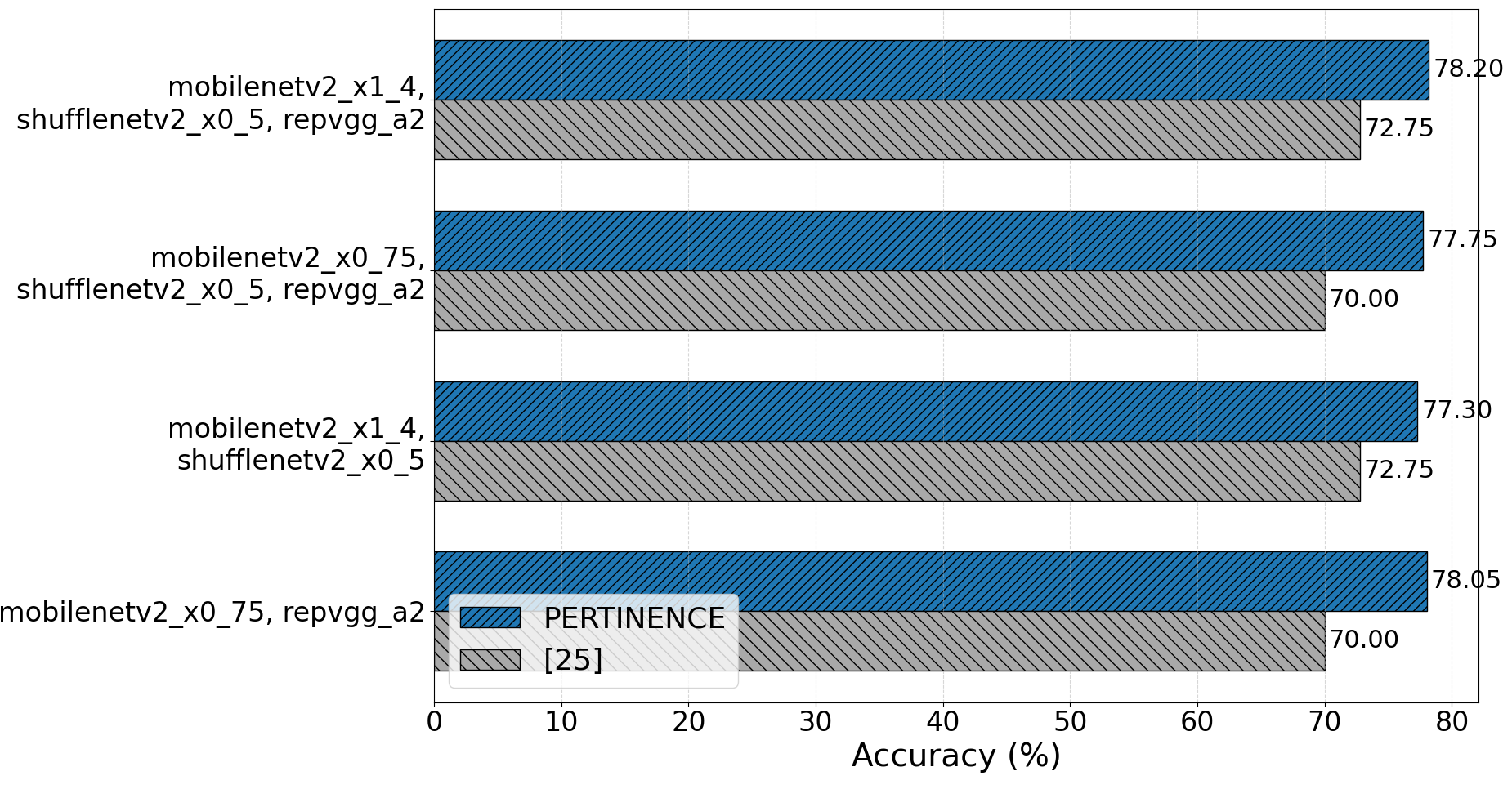}
    \caption{Comparison of Accuracy achieved by PERTINENCE and ~\cite{run-adapt1}}
    \label{fig:rel_work}
\end{figure}

As shown in figure~\ref{fig:rel_work}, PERTINENCE consistently outperforms the adaptive model-selection approach of~\cite{run-adapt1}, achieving 5–8\% higher accuracy across multiple model combinations. For instance, with,  \texttt{mobilenetv2\_x1\_4}, \texttt{shufflenetv2\_x0\_5}, \texttt{repvgg\_a2} PERTINENCE reaches 78.2\% accuracy, compared to 72.75\% achieved  by \cite{run-adapt1}. 
Moreover, the approach in \cite{run-adapt1} trains an offline predictive model to select the fastest DNN that correctly classifies a given input, based on handcrafted image features as mentioned earlier. While this yields low computational cost, it is highly dependent on feature engineering. In contrast, PERTINENCE leverages learned feature representations from pre-trained SOTA backbones to make dispatching decisions directly from data, eliminating the need for manual feature design.
Moreover, as opposed to~\cite{run-adapt1} offering a single final solution, PERTINENCE explores trade-offs among different attributes (such as accuracy, MFLOPs, power, inference time), offering multiple solutions to choose from.
Although~\cite{run-adapt1} achieves slightly lower computational overhead, PERTINENCE delivers superior accuracy, adaptability, and control over computational cost–performance tradeoff, making it more practical for real-time and resource-constrained scenarios.

\label{sec: alt_methods}
Furthermore, before moving to an NN-based solution for the Input Dispatcher, we also explored feature-engineering-driven dispatching methods similar to those in~\cite{run-adapt1}. Specifically, we employed algorithms like Histogram of Gradients (HoG), Local Binary Patterns (LBP), and BRISQUE, alongside metrics such as the Structural Index (SI) and Gradient Structural Index (GSI). These features were fed into lightweight classifiers such as Support Vector Machines (SVM), random forests (RF) and gradient boosting to predict the most appropriate DNN for each input. However, these traditional approaches consistently produced lower dispatch accuracy compared to NN-based methods. 
Finally, to the best of our knowledge, the proposed work is the first to explore dynamic input dispatching for large DNN models, such as Vision Transformers (ViTs).

\bgroup\color{black}
\subsection{Comparison with Early-Exit, Slimmable networks Based Baselines }

\begin{figure}[b]
    \centering
   \includegraphics[width=\columnwidth]{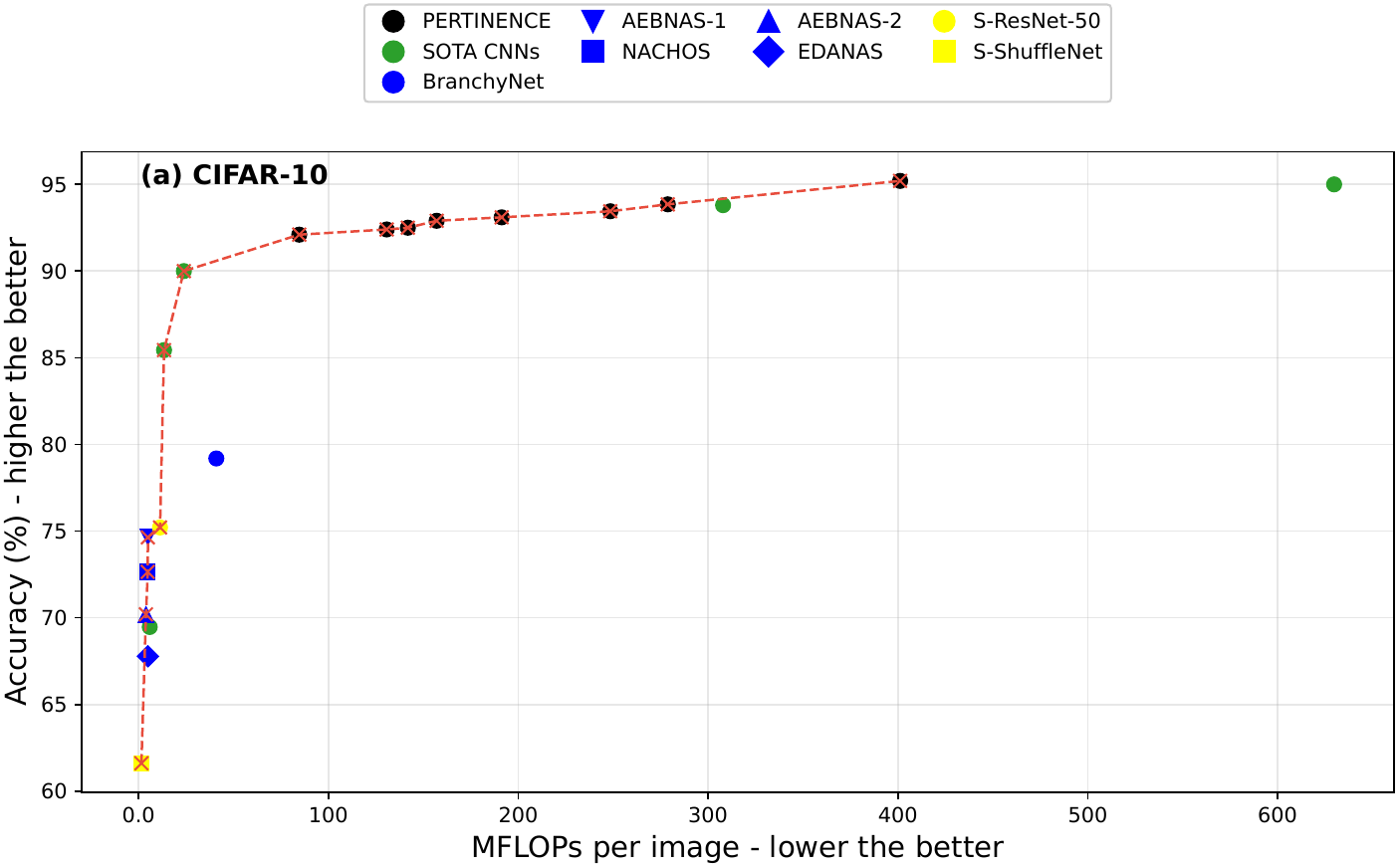}
   \caption{Accuracy vs MFLOPs for PERTINENCE and Early-Exit, Slimmable networks Based Baselines - CIFAR-10}
   \label{fig:rel_work_ee_slim-cifar10}
\end{figure}

In this subsection, we compare PERTINENCE against a set of adaptive inference baselines, including AlexNet-BranchyNet~\cite{branchynet}, slimmable networks (S-ResNet-50, S-ShuffleNet) ~\cite{slim1, slim2}, and NAS-based early-exit methods comprising EDANAS~\cite{ednas}, NACHOS~\cite{nachos}, and AEBNAS~\cite{aebnas}. 
Figure~\ref{fig:rel_work_ee_slim-cifar10} presents the accuracy versus average MFLOPs trade-offs per image for the CIFAR-10 dataset for the mentioned approaches and PERTINENCE.
PERTINENCE forms a well-defined Pareto front across eight operating configurations, spanning 92.1\%-95.2\% accuracy at 84.65--401.17 MFLOPs per image. At its peak configuration, PERTINENCE achieves 95.2\% accuracy at 401.17 MFLOPs, surpassing SOTA-vgg16\_bn (95.0\% at 629.76 MFLOPs), delivering superior accuracy at a \texttt{36.3\%} reduction in computational cost. At its lightest configuration, PERTINENCE achieves 92.1\% at 84.65 MFLOPs, a region where no competing adaptive baseline approaches this level of accuracy. 
The entire PERTINENCE Pareto front operates above 92\% accuracy, whereas the most accurate existing adaptive method, AlexNet-BranchyNet~\cite{branchynet}, reaches only 79.19\% at 40.98 MFLOPs, a gap of nearly 13 percentage points at a similar MFLOPs value. As for Slimmable baselines~\cite{slim1, slim2}, S-ResNet-50 achieves 75.21\% accuracy at 111.34 MFLOPs, and S-ShuffleNet achieves 61.61\% at 1.52 MFLOPs. NAS-based early-exit methods, despite hardware-aware architecture search, remain confined to low accuracy regimes: MobileNetV3-based AEBNAS~\cite{aebnas} provides two solutions: AEBNAS-1 achieves 74.64\% at 4.94 MFLOPs and AEBNAS-2 achieves 70.20\% at 3.92 MFLOPs, NACHOS~\cite{nachos} achieves 72.65\% at 4.88 MFLOPs, and EDANAS~\cite{ednas} achieves 67.78\% at 4.94 MFLOPs. While these methods operate at substantially lower MFLOPs, which allows them to enter the Pareto front, they offer lower accuracy than PERTINENCE. Moreover, PERTINENCE offers a flexible approach able to find different trade-offs between accuracy and MFLOPs, which can be adapted to different scenarios.

Figure~\ref{fig:rel_work_ee_slim-cifar100} presents the corresponding comparison on CIFAR-100. PERTINENCE spans an accuracy range of 76.25\%-77.3\% at 269.93-324.48 MFLOPs per image, a region of the Pareto space that no competing baseline occupies. The closest competing methods in accuracy are the slimmable pruned ResNet-50 variants~\cite{slimprun}: ResNet-50$\times$0.75 achieves 78.1\% (0.8\% improvement) but at 2300 MFLOPs, and ResNet-50 $\times$0.5 achieves 77.7\% (0.4\% improvement) at 1100 MFLOPs representing compute overheads of 7.1$\times$ and 3.4$\times$ respectively relative to PERTINENCE for comparable accuracy. 
SOTA-repvgg\_a1 achieves 77.65\% but at 1715.7 MFLOPs, 5.3$\times$
more compute than PERTINENCE's top configuration. Below the PERTINENCE operating range, the universally slimmable MobileNetV2 (US-MobileNetV2)~\cite{slim1} achieves 71.61\% at 176.22 MFLOPs, and MobileNetV3-based AEBNAS~\cite{aebnas} achieves 70.12\% at 33.34 MFLOPs and 69.9\% at 29.6 MFLOPs, operating at lower MFLOPs but with accuracy deficits of 5-7 percentage points relative to the lowest PERTINENCE configuration. These results demonstrate that PERTINENCE achieves accuracy competitive with the most computationally expensive slimmable pruned baselines at a fraction of their MFLOPs cost, while simultaneously exceeding the accuracy of all lightweight early-exit and slimmable configurations by a substantial margin.

\begin{figure}[t]
    \centering
    \includegraphics[width=\columnwidth]{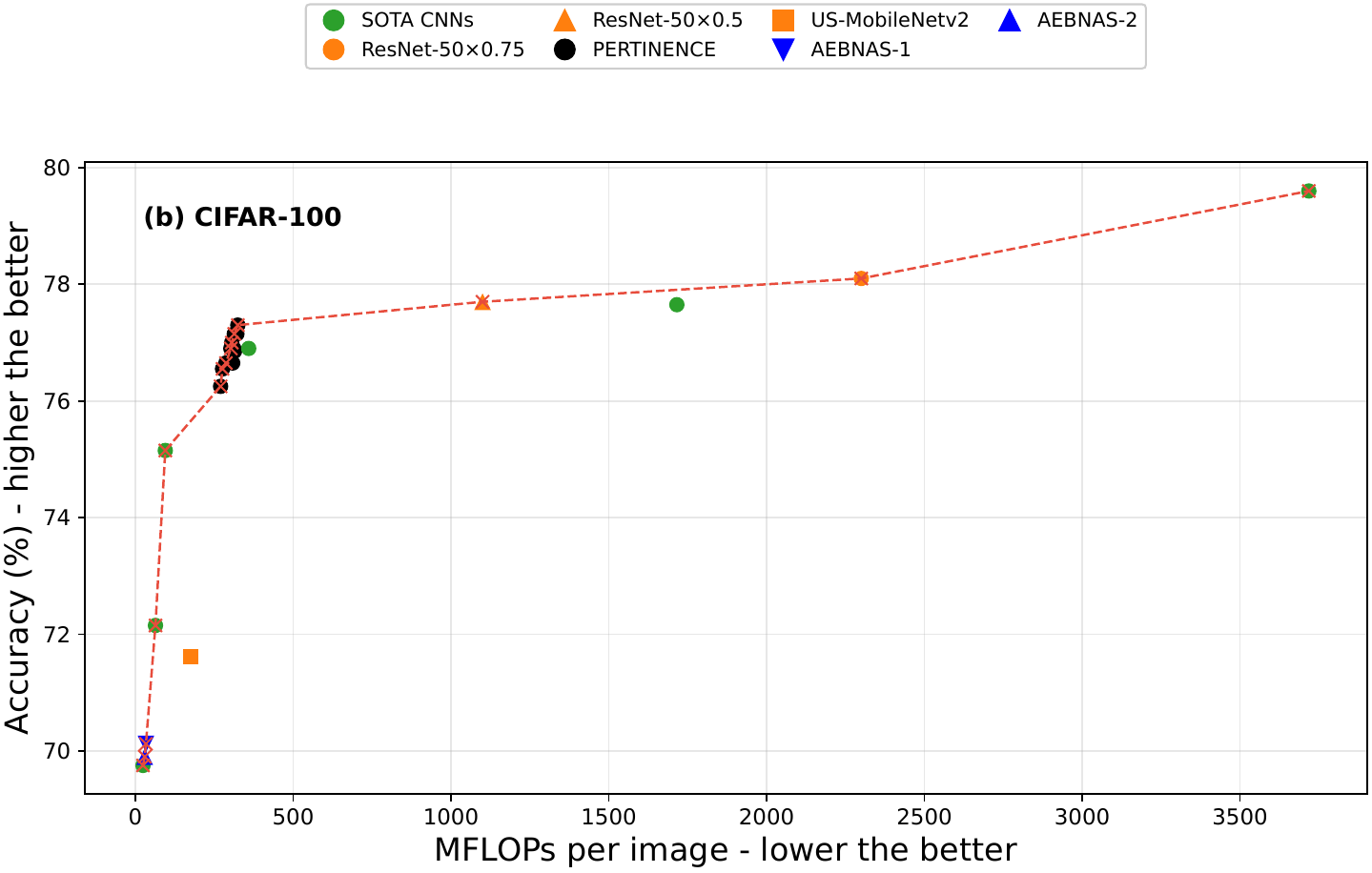}
    \caption{Accuracy vs MFLOPs for PERTINENCE and Early-Exit, Slimmable networks Based Baselines - CIFAR-100}
    \label{fig:rel_work_ee_slim-cifar100}
\end{figure}
\egroup

\section{Conclusion}
In this paper, we propose PERTINENCE, an online method for DNN-based tasks that dynamically dispatches inputs to the most lightweight model capable of correctly processing them.
Firstly, PERTINENCE is useful for ML users to dynamically combine existing pre-trained models to obtain \textit{alternative choices} -- in terms of the trade-off between accuracy and complexity -- without the cumbersome task of creating new models from scratch or undergoing challenging training procedures.
We explored the space defined by the trade-off between accuracy and number of operations and find solutions where PERTINENCE could also dominate state-of-the-art DNN models, enabling improved accuracy with up to 36\% fewer operations thanks to the online opportunistic input dispatcher.
In future work, we aim to explore the PERTINENCE approach applied to other machine learning tasks and different input dispatcher approaches to further improve energy efficiency and accuracy.

\section*{Acknowledgment}
Part of this work was supported by Inria through the AxTRADE associate team and by the French National Research Agency (ANR) through the RADYAL project ANR-23-IAS3-0002 and REAxION project ANR-25-CE25-5926.
We acknowledge that we utilized artificial intelligence for grammar checking.

\bibliographystyle{ieeetr} 
\bibliography{ref.bib} 

\begin{IEEEbiography}[{\includegraphics[width=1in,height=1.25in,clip,keepaspectratio]{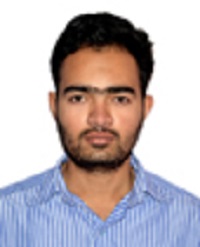}}]{Omkar Shende} is currently a Ph.D. student in the Department of Computer Science and Engineering at the Indian Institute of Technology Dharwad, India. He received his Master's degree in Advanced Computing from the Maulana Azad National Institute of Technology (MANIT), Bhopal in 2021. His research interests include edge AI systems, scheduling and fault-tolerant deep learning.
\end{IEEEbiography}
\vspace{-30\baselineskip}
\begin{IEEEbiography}[{\includegraphics[width=1in,height=1.25in,clip,keepaspectratio]{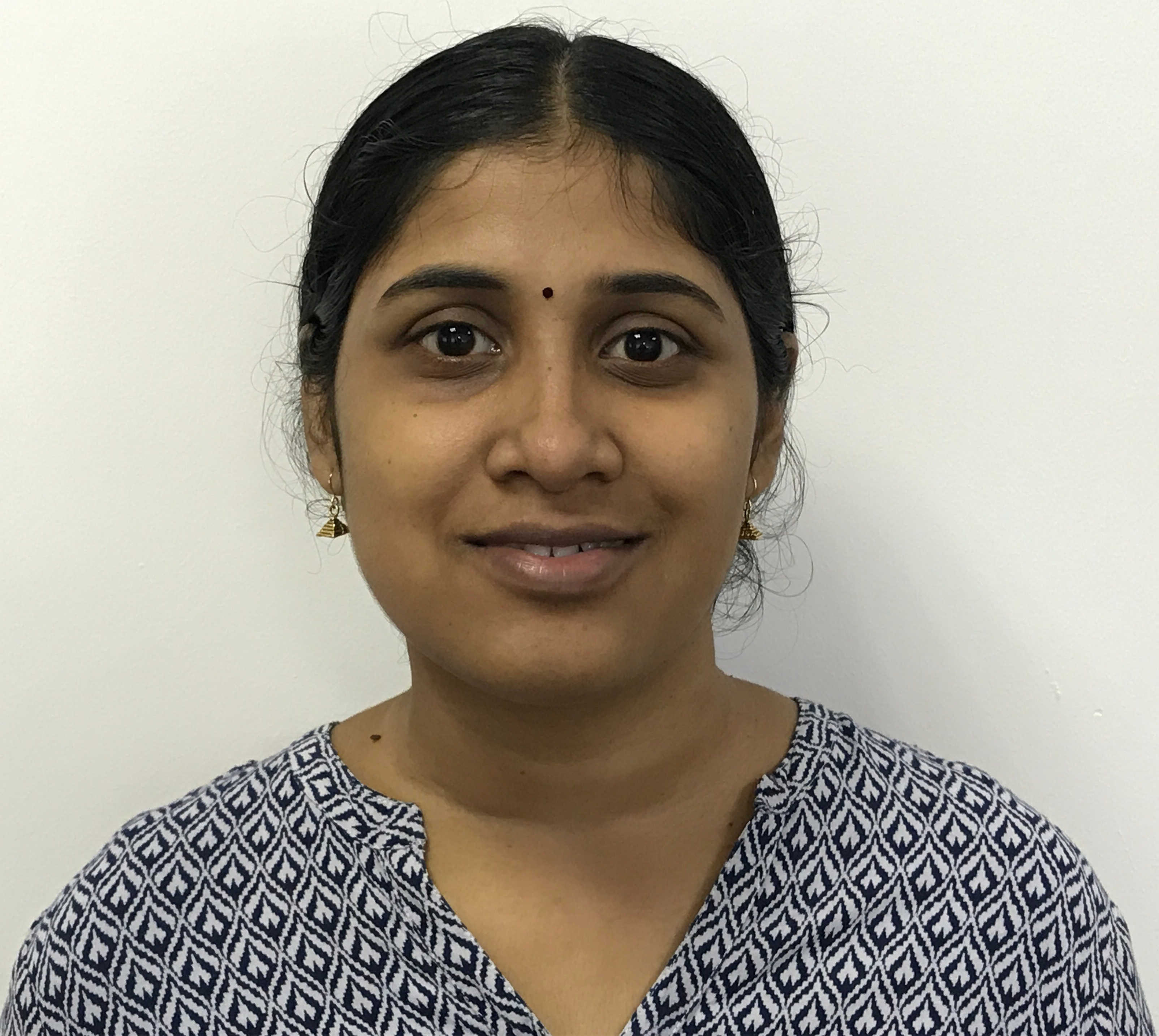}}]{Gayathri Ananthanarayanan} received the Ph.D. degree from the Indian
Institute of Technology Delhi, New Delhi, India, in
2017 and a master’s degree in embedded
systems in 2010 from the Birla Institute of Technology and
Science, Pilani, Pilani, India. Currently, she is an Assistant Professor
with the Department of Computer Science and Engineering, Indian Institute of
Technology Dharwad, Karnataka, India. Her current research interests include
computer architecture and embedded systems, with specific focus on hardware Performance analysis and efficient deployment of Edge-AI applications on Heterogeneous Multi-processor Platforms (HMPSoCs).
\end{IEEEbiography}
\vspace{-30\baselineskip}
\begin{IEEEbiography}[{\includegraphics[width=1in,height=1.25in,clip,keepaspectratio]{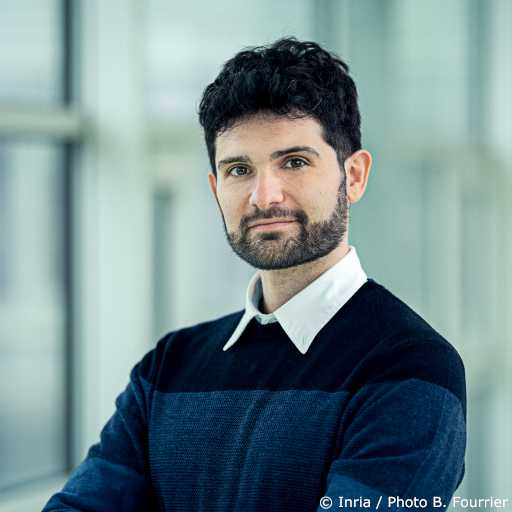}}]{Marcello Traiola}  is a tenured researcher at Inria (IRISA, Rennes, France), member of the TARAN team. He received a PhD in Computer Engineering from the University of Montpellier (2019) and an MSc (Laurea) in Computer Engineering from the University of Naples Federico II (2016). His research focuses on embedded systems, hardware accelerators, energy-efficient and approximate computing, with an emphasis on efficient design, reliability, and testing.
\end{IEEEbiography}

\end{document}

%% file: pareto-sota-dnns.tex
\begin{figure}[tp]
    \begin{subfigure}{.5\textwidth}
    \centering
    \includegraphics[width=.9\columnwidth]{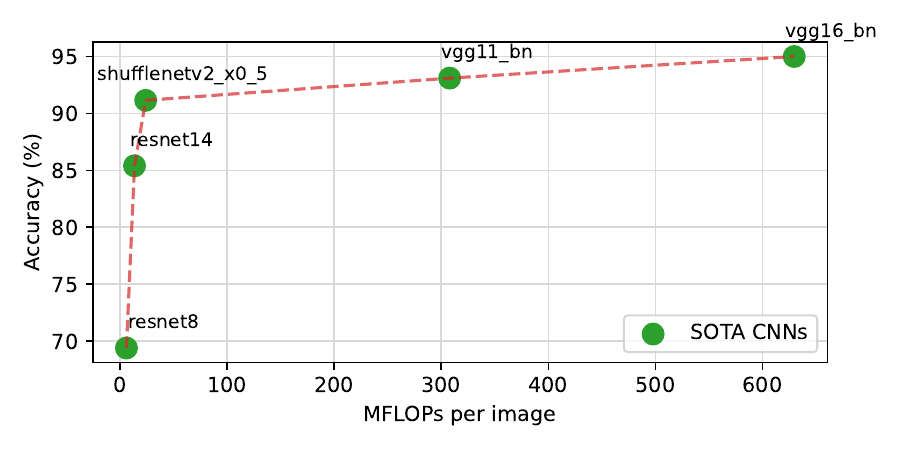}
    \caption{CIFAR-10\label{fig:cf10-pareto}}
    \end{subfigure}
    \begin{subfigure}{.5\textwidth}
    \centering
    \includegraphics[width=.9\columnwidth]{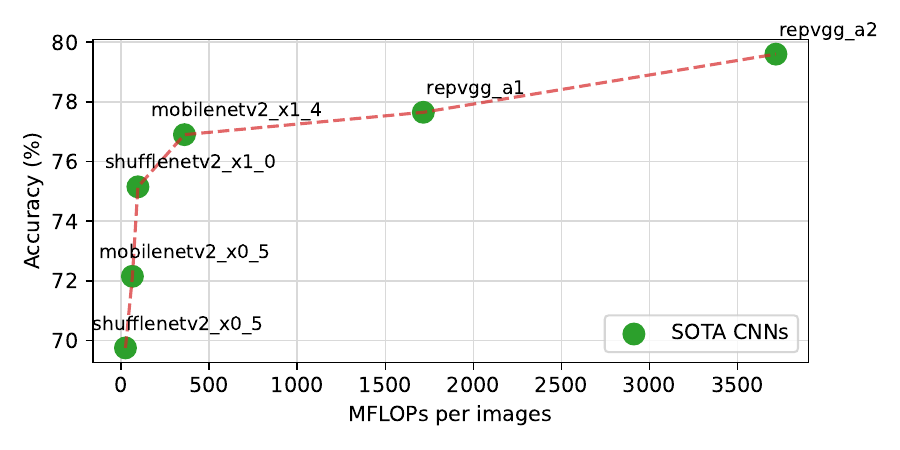}
    \caption{CIFAR-100\label{fig:cf100-pareto}}
    \end{subfigure}
     \begin{subfigure}{.5\textwidth}
    \centering
   \includegraphics[width=.9\columnwidth]{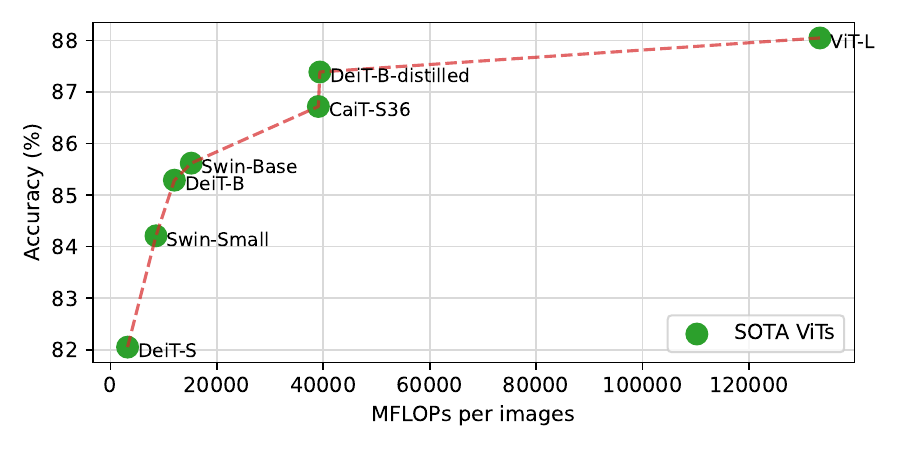}
    \caption{Tiny Imagenet with Vision transformers\label{fig:tinyimagenet-vit-pareto}}
    \end{subfigure}
    \caption{Pareto Front of DNN models trained on (a) CIFAR-10 (CNNs), (b) CIFAR-100 (CNNs), and (c) Tiny Imagenet (ViTs) datasets.}
    \label{fig:DNN-pareto}
\end{figure}

%% file: cifar10-Resnet8-feature-extract.tex
\begin{figure*}[t]
\centering 
\begin{subfigure}{0.333\textwidth}
\captionsetup{width=.9\linewidth}
  \includegraphics[width=\linewidth]{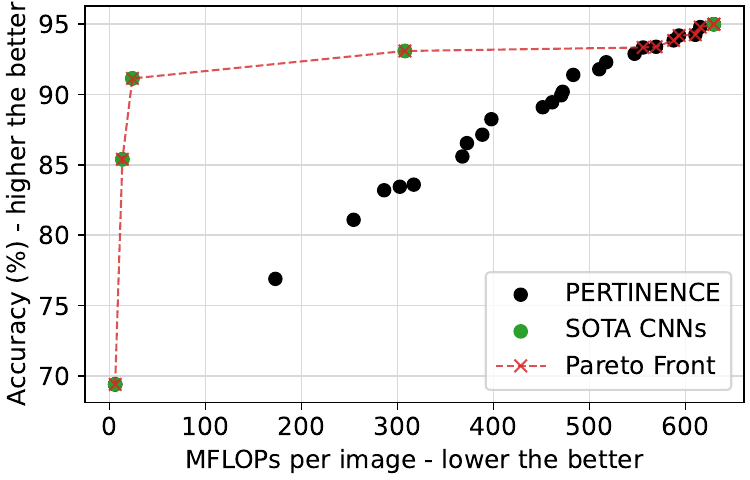}
\caption{Inputs dispatched to either \texttt{resnet8} or \texttt{vgg16\_bn} }
  \label{fig:q}
\end{subfigure}\hfil
\begin{subfigure}{0.333\textwidth}
\captionsetup{width=.9\linewidth}
\includegraphics[width=\linewidth]{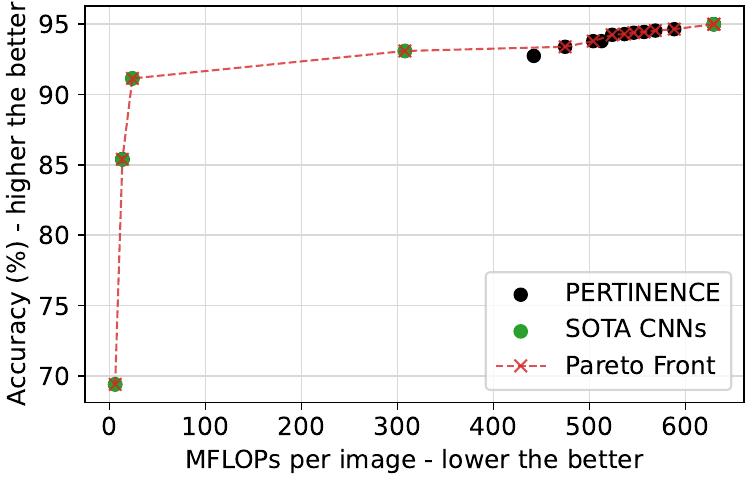}
\caption{Inputs dispatched to either \texttt{resnet14} or \texttt{vgg16\_bn}}
  \label{fig:b}
\end{subfigure}\hfil
\begin{subfigure}{0.333\textwidth}
\captionsetup{width=.9\linewidth}
\includegraphics[width=\linewidth]{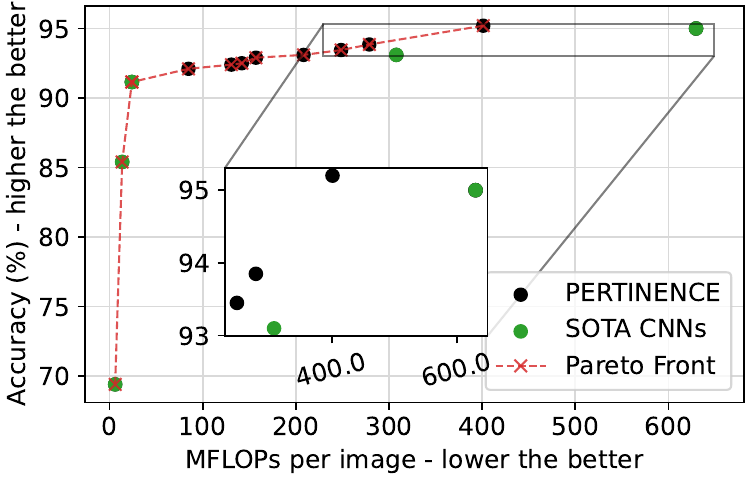}
\caption{Inputs dispatched to either \texttt{shufflenetv2\_x0\_5} or \texttt{vgg16\_bn}}
  \label{fig:cifar-10-res8-c}
\end{subfigure}\hfil
\begin{subfigure}{0.33\textwidth}
\captionsetup{width=.9\linewidth}
\includegraphics[width=\linewidth]{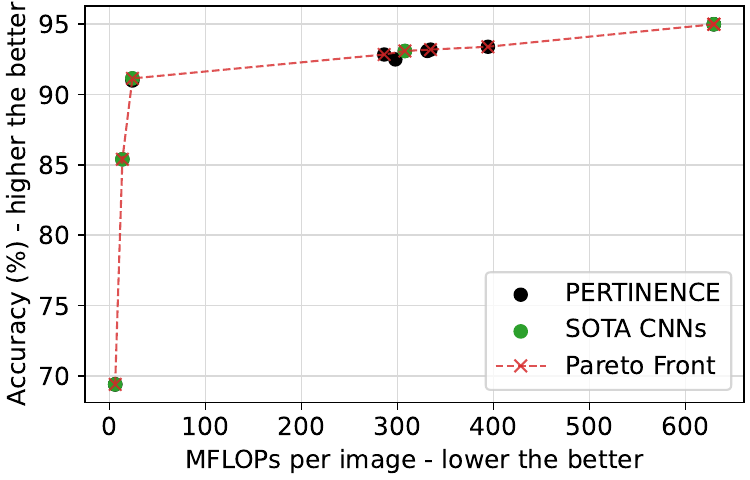}
\caption{Inputs dispatched to either \texttt{resnet8}, \texttt{shufflenetv2\_x0\_5}, or \texttt{vgg16\_bn}}
  \label{fig:c1}
\end{subfigure}\hfil
\begin{subfigure}{0.33\textwidth}
\captionsetup{width=.9\linewidth}
  \includegraphics[width=\linewidth]{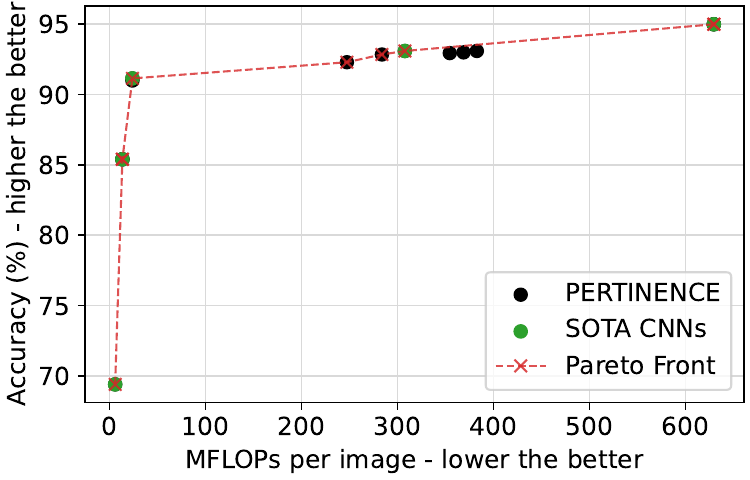}
\caption{Inputs dispatched to either \texttt{resnet14}, \texttt{shufflenetv2\_x0\_5}, or \texttt{vgg16\_bn}}
  \label{fig:d}
\end{subfigure}\hfil
\begin{subfigure}{0.33\textwidth}
  \includegraphics[width=\linewidth]{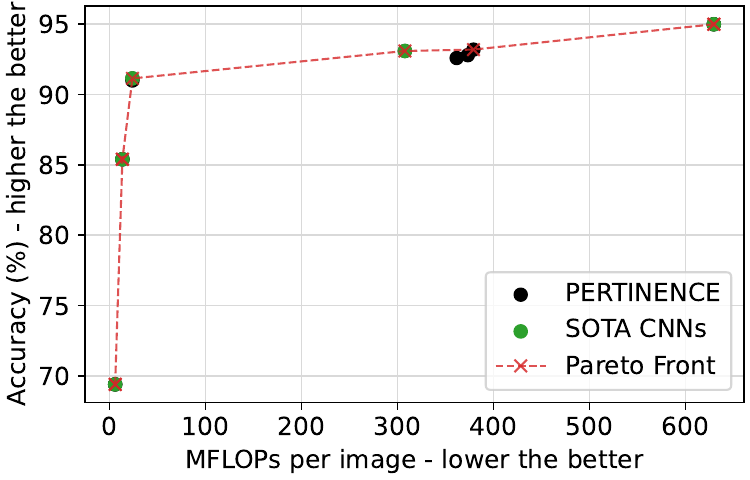}
\caption{\footnotesize Inputs dispatched to either \texttt{resnet8}, \texttt{resnet14}, \texttt{shufflenetv2\_x05}, or \texttt{vgg16\_bn}}
  \label{fig:e}
\end{subfigure}
\caption{Solutions obtained using \texttt{resnet8} feature extraction
capabilities on CIFAR-10 dataset. Subfigure~\ref{fig:cifar-10-res8-c} highlights the configurations where PERTINENCE dominates SOTA models.} 
\label{fig:cf10-res}
\end{figure*}

%% file: cifar10-Resnet14-feature-extract.tex
\begin{figure}[h]
\centering 
\begin{subfigure}{.85\columnwidth}
  \includegraphics[width=\linewidth]{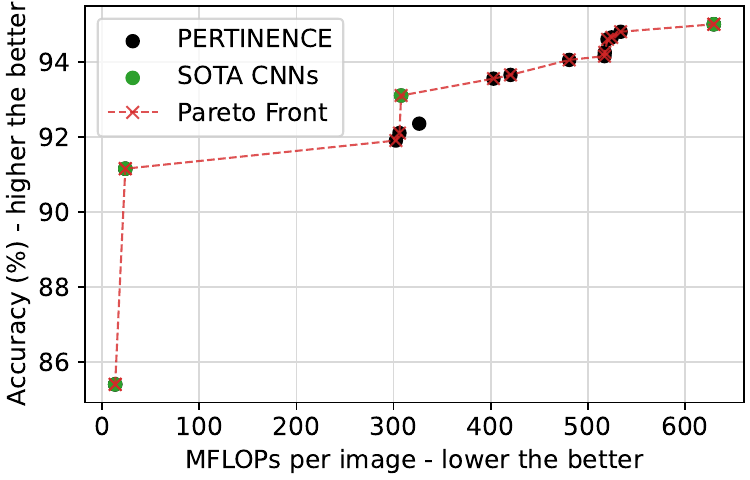}
  \caption{Inputs dispatched to either \texttt{resnet14}, or \texttt{vgg16\_bn}}
  \label{fig:a1}
\end{subfigure}\hfil
\begin{subfigure}{.85\columnwidth}
  \includegraphics[width=\linewidth]{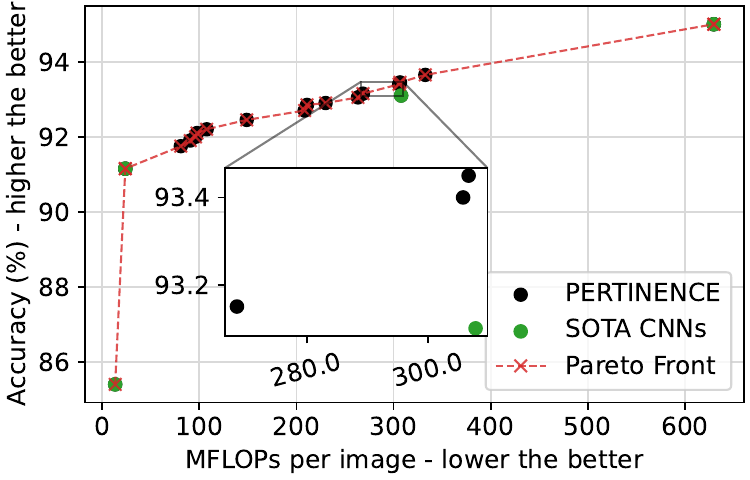}
  \caption{Inputs dispatched to either \texttt{shufflenetv2\_x0\_5}, or \texttt{vgg16\_bn}}
  \label{fig:b1}
\end{subfigure}\hfil
\begin{subfigure}{.85\columnwidth}
  \includegraphics[width=\linewidth]{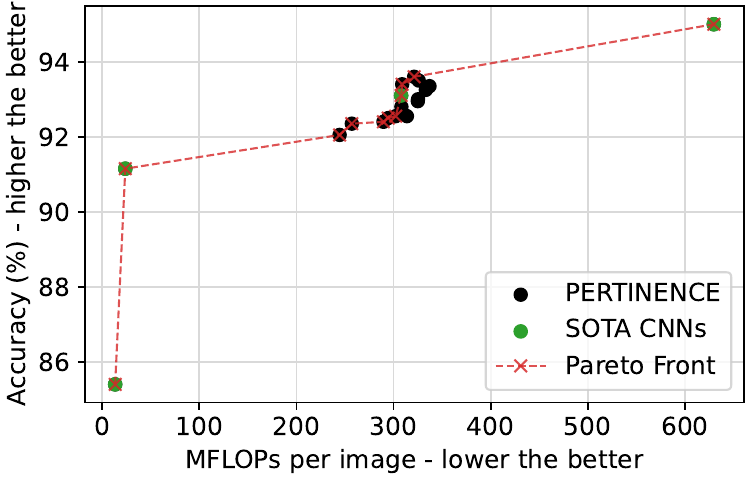}
  \caption{Inputs dispatched to either \texttt{resnet14}, \texttt{shufflenetv2\_x0\_5}, or \texttt{vgg16\_bn}}
  \label{fig:c}
\end{subfigure}
\vspace{-3pt}
\caption{Solutions obtained using \texttt{resnet14} feature extraction capabilities on CIFAR-10 dataset and CNNs. Subfigure~\ref{fig:b1}     highlights the configurations where PERTINENCE dominates SOTA models. }
\label{fig:cf10-res-r14f}
\end{figure}

%% file: cifar100-shufflenet-feature-extract.tex
\begin{figure}[p]
\centering 
\begin{subfigure}{.85\columnwidth}
  \includegraphics[width=\linewidth]{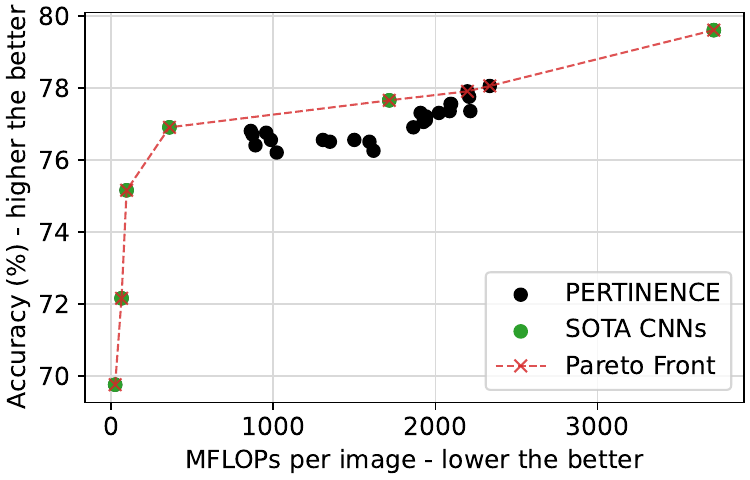}
\caption{Inputs dispatched to either 
\texttt{mobilenetv2\_x0\_75} or \texttt{repvgg\_a2}}
  \label{fig:b}
\end{subfigure}\hfil
\begin{subfigure}{.85\columnwidth}
\captionsetup{width=\linewidth}
  \includegraphics[width=\linewidth]{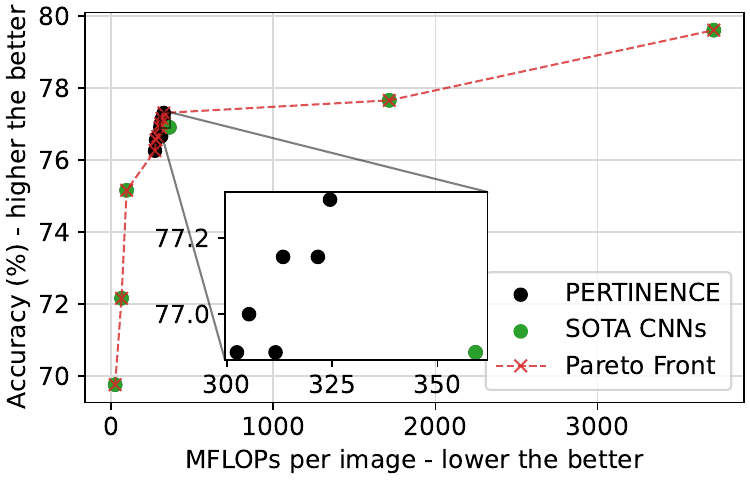}
\caption{Inputs dispatched to either \texttt{shufflenetv2\_x0\_5} or \texttt{mobilenetv2\_x1\_4}}
  \label{fig:c}
\end{subfigure}
\begin{subfigure}{.85\columnwidth}
  \includegraphics[width=\linewidth]{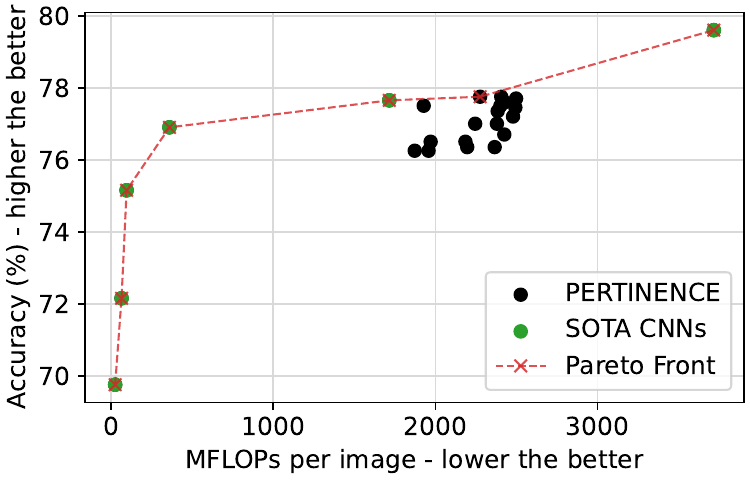}
\caption{Inputs dispatched to either \texttt{shufflenetv2\_x0\_5}, \texttt{mobilenetv2\_x0\_75}, or \texttt{repvgg\_a2}}
  \label{fig:d}
\end{subfigure}\hfil
\begin{subfigure}{.85\columnwidth}
  \includegraphics[width=\linewidth]{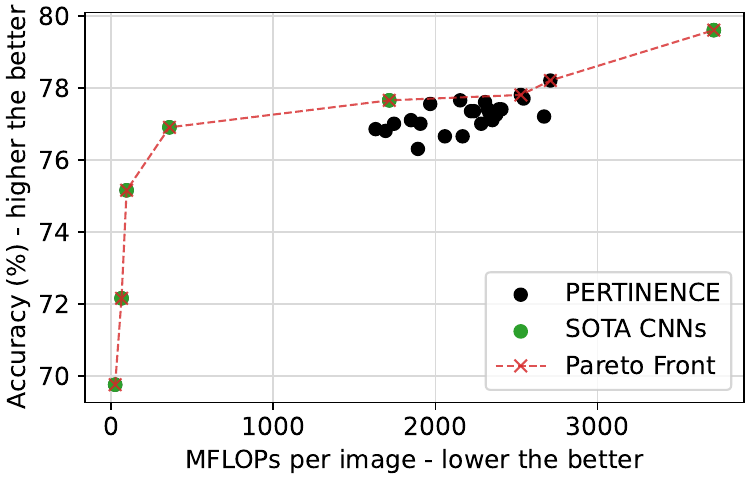}
\caption{Inputs dispatched to either \texttt{shufflenetv2\_x0\_5}, \texttt{mobilenetv2\_x1\_4}, or \texttt{repvgg\_a2}}
  \label{fig:d}
\end{subfigure}\hfil
\caption{Solutions obtained using \texttt{shufflenetv2\_x0\_5} feature extraction capabilities on CIFAR-100 dataset and CNNs.}
\label{fig:cf100-res}
\end{figure}

%% file: tinyimagenet-resnet50-feature-extract.tex
\begin{figure}[!h]
\centering 
\begin{subfigure}{.9\columnwidth}
  \includegraphics[width=\linewidth]{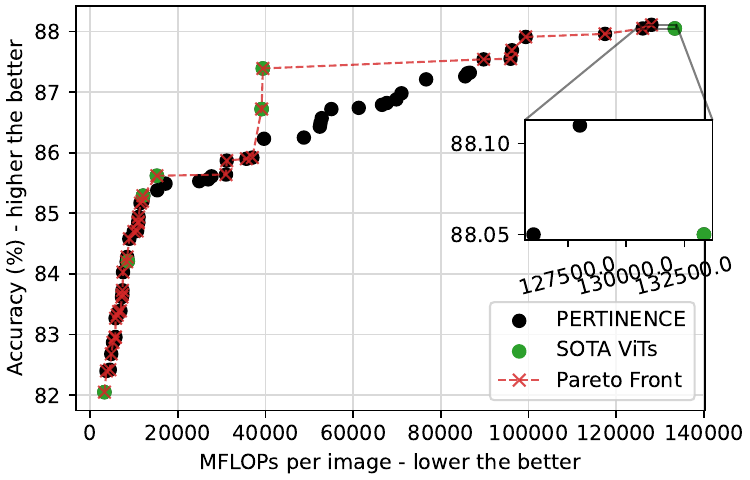}
\caption{Inputs dispatched to either 
\texttt{DeiT-S}, \texttt{DeiT-B} or \texttt{ViT-L}}
  \label{fig:tinyimagenetA}
\end{subfigure}\hfil
\begin{subfigure}{.9\columnwidth}
\captionsetup{width=\linewidth}
  \includegraphics[width=\linewidth]{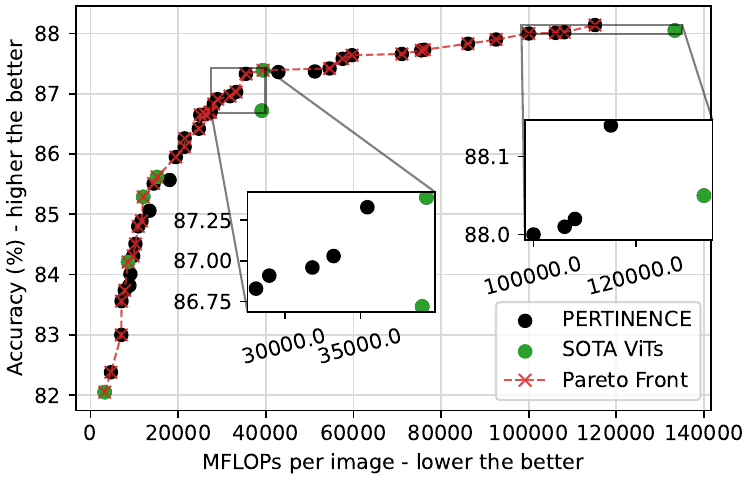}
\caption{Inputs dispatched to either 
\texttt{DeiT-S}, \texttt{Swin Base}, \texttt{DeiT-distilled}, or \texttt{ViT-L}}
  \label{fig:tinyimagenetB}
\end{subfigure}

\caption{Solutions obtained using \texttt{resnet50} feature extraction capabilities on TinyImagenet dataset and Vision Transformers.}
\label{fig:tinyimagenet-res}
\end{figure}

%% file: power.tex
\begin{figure}[!h]
\centering 
\begin{subfigure}{.9\columnwidth}
  \includegraphics[width=\linewidth]{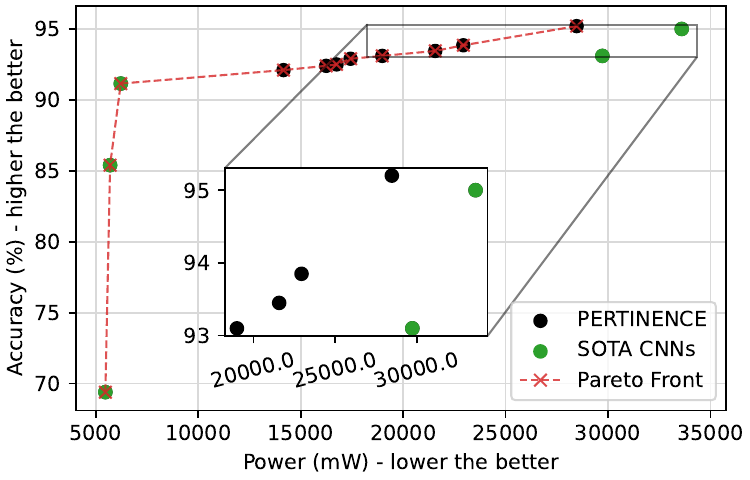}
\caption{Feature extracted through \texttt{resnet8} and inputs dispatched to either \texttt{shufflenetv2\_x0\_5} or \texttt{vgg16\_bn}. Compare to Figure~\ref{fig:cifar-10-res8-c}}
  \label{fig:power6c}
\end{subfigure}\hfil
\begin{subfigure}{.9\columnwidth}
  \includegraphics[width=\linewidth]{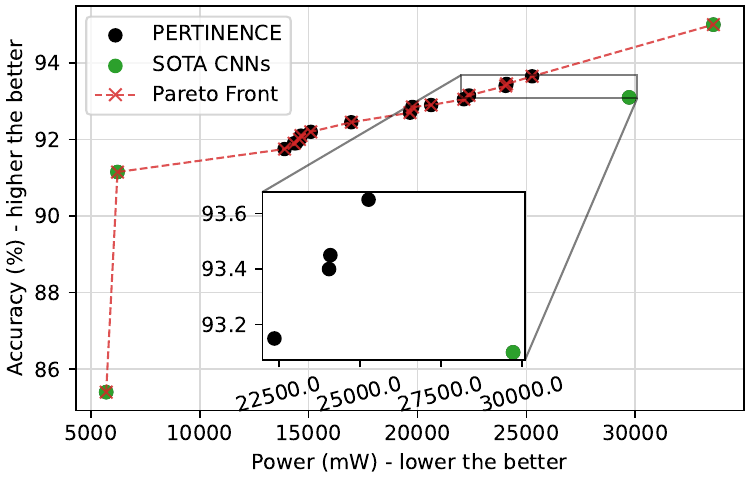}
\caption{Feature extracted through \texttt{resnet14} and inputs dispatched to either \texttt{shufflenetv2\_x0\_5} or \texttt{vgg16\_bn}. Compare to Figure~\ref{fig:cf10-res-r14f}b}
  \label{fig:power7b}
\end{subfigure}\hfil
\caption{Trade off between power consumption and accuracy for some explorations for the CIFAR-10 dataset.}
\label{fig:power}
\end{figure}